\documentclass[10pt,twocolumn,letterpaper]{article}

\usepackage{iccv}
\usepackage{times}
\usepackage{epsfig}
\usepackage{graphicx}
\usepackage{amsmath}
\usepackage{amssymb}

\usepackage{booktabs}
\usepackage[font={small}]{caption}
\usepackage[labelformat=simple]{subcaption}

\usepackage{color, colortbl}
\definecolor{Gray}{gray}{.9}
\newcolumntype{g}{>{\columncolor{Gray}}c}

\usepackage[pagebackref=true,breaklinks=true,letterpaper=true,colorlinks,bookmarks=false]{hyperref}

\iccvfinalcopy 

\begin{document}

\title{TSP: Temporally-Sensitive Pretraining of Video Encoders for Localization Tasks} 

\author{
Humam Alwassel \quad
Silvio Giancola \quad
Bernard Ghanem \\
King Abdullah University of Science and Technology (KAUST) \\
{\tt\small\{humam.alwassel,silvio.giancola,bernard.ghanem\}@kaust.edu.sa} \\
{{{\url{http://humamalwassel.com/publication/tsp}}}}
}

\maketitle

\newcommand{\ahat}{\hat{\textbf{a}}}
\newcommand{\av}{\textbf{a}}
\newcommand{\bv}{\textbf{b}}
\newcommand{\cv}{\textbf{c}}
\newcommand{\dv}{\textbf{d}}
\newcommand{\uv}{\textbf{u}}
\newcommand{\vv}{\textbf{v}}
\newcommand{\x}{\textbf{x}}
\newcommand{\X}{\textbf{X}}
\newcommand{\y}{\textbf{y}}
\newcommand{\Y}{\textbf{Y}}
\newcommand{\z}{\textbf{z}}
\newcommand{\w}{\textbf{w}}
\newcommand{\W}{\textbf{W}}
\newcommand{\p}{\textbf{p}}
\newcommand{\q}{\textbf{q}}
\newcommand{\h}{\textbf{h}}
\newcommand{\A}{\textbf{A}}
\newcommand{\C}{\textbf{C}}
\newcommand{\D}{\textbf{D}}
\newcommand{\F}{\textbf{F}}
\newcommand{\V}{\textbf{V}}
\newcommand{\U}{\textbf{U}}
\newcommand{\I}{\textbf{I}}
\newcommand{\PX}{\textbf{P}}
\newcommand{\mSigma}{\mathbf{\Sigma}}
\newcommand{\0}{\mathbf{0}}
\newcommand{\1}{\mathbf{1}}

\begin{abstract}
Due to the large memory footprint of untrimmed videos, current state-of-the-art video localization methods operate atop precomputed video clip features. These features are extracted from video encoders typically trained for trimmed action classification tasks, making such features not necessarily suitable for temporal localization. In this work, we propose a novel supervised pretraining paradigm for clip features that not only trains to classify activities but also considers background clips and global video information to improve temporal sensitivity. Extensive experiments show that using features trained with our novel pretraining strategy significantly improves the performance of recent state-of-the-art methods on three tasks: Temporal Action Localization, Action Proposal Generation, and Dense Video Captioning. We also show that our pretraining approach is effective across three encoder architectures and two pretraining datasets. We believe video feature encoding is an important building block for localization algorithms, and extracting temporally-sensitive features should be of paramount importance in building more accurate models.
The code and pretrained models are available on our \href{http://humamalwassel.com/publication/tsp/}{project website}.
\vspace{-10pt}
\end{abstract}
\section{Introduction}\label{sec:introduction}

\begin{figure}[t!]
    \centering
    \includegraphics[width=\linewidth]{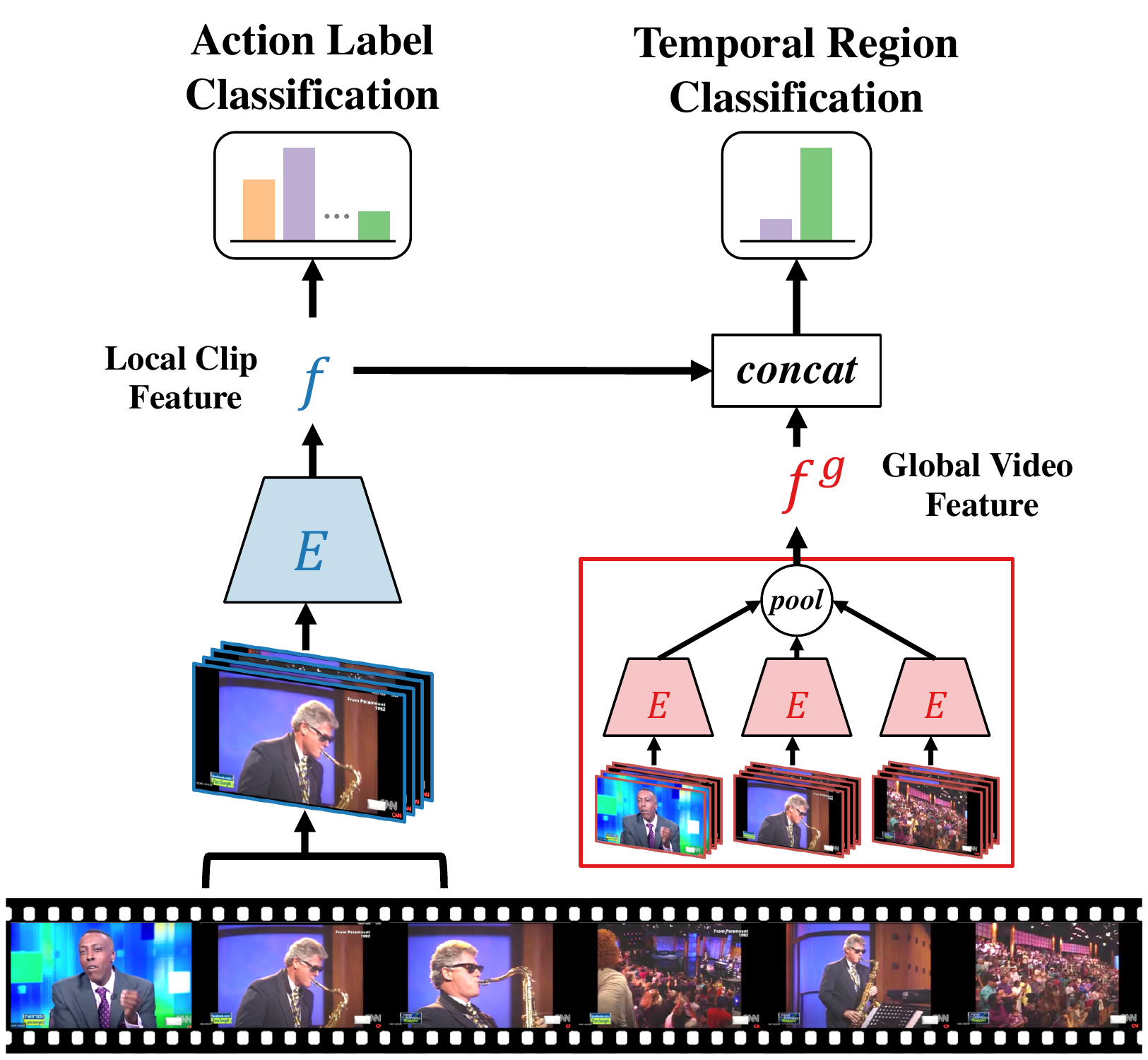}
    \vspace{-15pt}
    \caption{\textbf{Temporally-Sensitive Pretraining (TSP)}. 
    We train video encoders to be temporally-sensitive through a novel supervised pretraining paradigm. A fixed-sized clip is sampled from an untrimmed video and passed through the encoder to obtain a local clip feature (blue). A global video feature (red) is pooled from the local features of all clips in the untrimmed video. The local and global features are used to train the encoder on the task of classifying the label of foreground clips (action label) and classifying whether a clip is inside or outside the action (temporal region).}
    \vspace{-3pt}
    \label{fig:pipeline}
\end{figure}

Video understanding is thriving in the computer vision community, and it manifests in several challenging tasks such as action classification~\cite{Feichtenhofer_2019_ICCV,Korbar_2019_ICCV,Lin_2019_ICCV_TSM,Tran_2019_ICCV}, activity localization~\cite{gao_eccv_2018,kumar2017hide,zhao2020bottom}, and video captioning~\cite{Hou_2019_ICCV,Pan_2020_CVPR_Spatio,Wang_2019_ICCV,Zheng_2020_CVPR_syntax}. Yet, the success of video research has been lagging behind that of its counterpart in the image domain. In many aspects, this is due to the exponentially larger amount of data in videos compared to images, not fitting in commodity hardware. Image encoders have the privilege to process batches of complete images at once, thus exploiting the rich contextual information from all pixels. Empowered by such capability, many image models are trained in an end-to-end manner for complex tasks such as object detection~\cite{redmon2016you,ren2015faster,tan2020efficientdet}, semantic segmentation~\cite{chen2018encoder,he2017mask,howard2019searching}, and image captioning~\cite{anderson2018bottom,you2016image,lu2017knowing}. In contrast, the long and variable length of \emph{untrimmed} videos makes it impractical to encode a complete video on current hardware accelerators~\cite{wu2019long}. 
While a few recent localization works~\cite{liu2020progressive,zhao_iccv_2017} attempt to train end-to-end for \emph{untrimmed} video tasks, such as temporal action localization, they need to resort to aggressive spatial and temporal downsampling to remain computationally practical.
Instead, most state-of-the-art localization methods for \emph{untrimmed} videos choose to learn models atop \emph{precomputed} clip features~\cite{bmt,Lin_2019_ICCV,xu2020gtad,Zeng_2019_ICCV}.

In this work, we focus on improving the precomputed features used for temporal localization tasks, which we define as tasks that require predictions related to the time dimension of the video. Specifically, we target three important localization problems: Temporal Action Localization (TAL), Action Proposal Generation (Proposals), and Dense Video Captioning (Dense-Captioning). State-of-the-art methods for these localization tasks use features extracted from video encoders typically pretrained for the task of \emph{Trimmed Action Classification} (TAC) on large-scale datasets, such as Kinetics~\cite{dataset_kinetics} and Sports-1M~\cite{dataset_sports1m}. However, this pretrained representation is not necessarily suitable for localization tasks. In particular, we observe that TAC-pretrained features tend to be temporally-insensitive, \ie background (no action) segments can have quite similar representations to foreground (action) segments from the same untrimmed video. We provide an analysis study of TAC-pretrained features in Section~\ref{sec:analysis} that shows evidence of the high cosine similarity between features of background and foreground clips. These temporally-insensitive features make it harder for the localization algorithm to learn the target task, and thus, negatively impact the final performance.

To circumvent these drawbacks, we propose a novel, supervised pretraining paradigm for video clip representation that not only trains to classify foreground activities but also considers background clips and global video information to improve temporal sensitivity. We refer to our pretraining approach as \emph{Temporally-Sensitive Pretraining} (TSP). Figure~\ref{fig:pipeline} gives an overview of TSP. We conduct extensive experiments to show that features extracted by clip encoders pretrained with TSP are more discriminative, and that training state-of-the-art localization algorithms atop TSP features results in significant performance gains on three temporal localization tasks: TAL, Proposals, and Dense-Captioning. Moreover, TSP gives consistent performance boosts regardless of the video encoder architecture, pretraining dataset, or the localization algorithm learned atop our features. Interestingly, we observe that localization performance on short instances greatly improves when using TSP pretrained features. This aligns well with our hypothesis that temporally-sensitive features allow localization algorithms to draw sharper contrast between foreground and background context in long untrimmed videos.

\vspace{3pt}\noindent\textbf{Contributions.}
\textbf{(I)} We propose TSP, a temporally-sensitive supervised pretraining task for video encoders. TSP trains an encoder to explicitly discriminate between foreground and background clips in untrimmed videos.
\textbf{(II)} We show with comprehensive experiments that using features pretrained with the TSP task significantly improves performance across three video localization problems. Additionally, we show the generalization capability of our pretraining strategy on three encoder architectures and two pretraining datasets. We also demonstrate consistent performance gains for multiple localization algorithms trained on the same target problem. 
\textbf{(III)} We provide an extensive analysis study of our features. Interestingly, we observe that TSP pretraining boosts temporal action localization performance on short action instances. The study also demonstrates that our features are in fact temporally-sensitive and can encode background clips differently from foreground clips.

\section{Related Work}\label{sec:related_work}

\noindent\textbf{Action recognition.}
Large-scale video datasets, such as UCF-101~\cite{dataset_ucf101}, Sports-1M~\cite{dataset_sports1m}, and Kinetics~\cite{dataset_kinetics}, have accelerated the development of action classification models. Simonyan and Zisserman~\cite{two_stream} introduced a two-stream encoder to represent appearance with RGB frames and motion with stacked optical flow vectors. Wang~\etal~\cite{tsn} proposed the Temporal Segment Network (TSN) encoder to capture long-term temporal information. Pretrained on TAC, TSN along with other recent architectures (\eg R(2+1)D~\cite{tran2018closer}, I3D~\cite{i3d}, and C3D~\cite{c3d}) have become the \emph{de facto} feature extractors for temporal action localization (TAL)~\cite{piergiovanni_cvpr_2018}, action segmentation~\cite{action_segmentation}, and event captioning~\cite{event_captioning}. Since TAC-pretraining is not necessarily suitable for these localization tasks, we propose a pretraining that learns from both foreground and background clips in untrimmed videos.

\vspace{3pt}\noindent\textbf{Temporal action localization and proposal generation.}
Many algorithms have been developed for TAL~\cite{alwassel_2018_actionsearch,dai_iccv_2017,activitynet_challenge,shou_cvpr_2017,yeung_cvpr_2016}. While the majority has been on fully-supervised TAL~\cite{TSA_Net,li2020deep,dbg,liu2020progressive,long2019gaussian}, recent works have also studied TAL under weak supervision~\cite{basnet_aaai20,liu2019weakly,pardo2021refineloc,paul_eccv_2018,shou_eccv_2018}, single-frame supervision~\cite{ma2020sfnet}, and self-supervision~\cite{Jain_2020_CVPR}. The first generation of algorithms applied complex action classifiers in a sliding window fashion~\cite{gaidon_ijcv_2013, oneata_cvpr_2014}. To alleviate the expensive cost of sliding an action classifier over long videos, the second generation of algorithms~\cite{buch_cvpr_2017, gao_iccv_2017,Lin_2019_ICCV,lin_eccv_2018,shou_cvpr_2016,Zeng_2019_ICCV} followed a two-stage approach that first learns action proposals to limit the number of candidates passed to the action classifier. A third set of algorithms jointly learn action proposals and action classifiers in one stage~\cite{chao_cvpr_2018,xu_iccv_2017,xu2020gtad,zhao_iccv_2017}. A few works~\cite{liu2020progressive,zhao_iccv_2017} learn TAL end-to-end by drastically downsampling videos to be computationally practical, \eg PBRNet~\cite{liu2020progressive} uses only 3 frames per second on ActivityNet and SSN~\cite{zhao_iccv_2017} uses only 9 clips per proposal. In contrast, most state-of-the-art methods build atop precomputed features from TAC-pretrained encoders. Since experiments show that such features are not best suited for TAL and Proposals, we propose to replace them with temporally-sensitive pretrained features that can significantly boost performance.

\vspace{3pt}\noindent\textbf{Dense video captioning.}
Krishna~\etal~\cite{activitynet_captions_dataset} introduced the task of Dense-Captioning along with the ActivityNet Captions benchmark. Dense-Captioning aims at both localizing and textually describing all events in a video. This problem branched out from video captioning~\cite{wang2018video,yan2019stat,pei2019memory}, where a full video is captioned without localizing events. \cite{activitynet_captions_dataset} uses a variant of DAPs~\cite{escorcia2016daps} to generate proposals and employs an LSTM-based captioning module to describe these proposals. Subsequent works use bidirectional attentive fusion~\cite{Wang_2018_CVPR}, masked transformers~\cite{Zhou_2018_CVPR}, and reinforcement  learning~\cite{Li_dvc,sdvc,mft}. A line of multi-modal Dense-Captioning methods combine visual cues with signals from audio~\cite{rahman2019watch}, speech/subtitles \cite{shi2019dense}, or both \cite{bmt,mdvc}. Similar to TAL and Proposals, Dense-Captioning algorithms rely on temporally-insensitive TAC-pretrained features, which do not perform as well as the TSP pretrained ones.

\section{Technical Approach}\label{sec:technical_approach}

\subsection{Traditional Pretraining Strategies} 
Since it is impractical to fit entire \emph{untrimmed} videos into commodity GPUs without drastically downsampling space or time, current state-of-the-art localization algorithms share a common practice in that they do not finetune their video encoders directly on the target task (\eg TAL). Instead, they use \textit{pretrained} encoders as \textit{fixed} feature extractors~\cite{bmt,Lin_2019_ICCV,xu2020gtad,Zeng_2019_ICCV}. Trimmed action classification (TAC) has been the traditional approach to pretrain these encoders. The TAC task aims to classify clips from short videos, where the action spans the entire video. While TAC has been successful in providing features that discriminate between different action classes, it often fails to distinguish between the action instance and its nearby background context. For example, recent diagnostic studies~\cite{alwassel_eccv_2018} have shown that state-of-the-art TAL methods are quite sensitive to the context around action instances and that their inability to distinguish between an action and its temporal background context is the main roadblock to improving localization performance. We argue that the features used in these state-of-the-art localization methods, pretrained on TAC, are a source of such confusion. Thus, we propose to depart from the traditional strategy and render the features temporally-sensitive through a novel pretraining task.

\subsection{How to Incorporate Temporal Sensitivity?}
A limiting  aspect of TAC-pretrained encoders is that they only learn from positive samples (foreground/action clips). Intuitively, learning from negative samples (background/no action clips) is expected to improve the temporal discriminative ability of these encoders. Given an untrimmed video, a good encoder for localization problems should be able to distinguish between the semantics of different actions as well as between actions and their background context.  Intuitively, clip features that have an idea of whether the clip is inside or outside an action can directly help localization methods find better activity/proposal boundaries for TAL and Proposals and find better captions for Dense-Captioning. Thus, we propose to pretrain encoders on the task of (1) classifying the label of foreground clips and (2) classifying whether a clip is inside or outside the action.

\subsection{Temporally-Sensitive Pretraining (TSP)}

\noindent\textbf{Input data.}
We pretrain our model using untrimmed videos with temporal annotations. The encoder is learned in an end-to-end fashion from the raw video input. In particular, given an untrimmed video, we sample a fixed-size input clip $\X$ of size $3$$\times$$L$$\times$$H$$\times$$W$, where $3$ refers to the RGB channels, $L$ is the number of frames, and $H$ and $W$ are the frame height and width. We assign $\X$ two labels: (1) the action class label $\y^c$ if this clip is from a foreground segment, and (2) the binary temporal region label $\y^r$ that indicates if the clip is from a foreground/action ($\y^r = 1$) or background/no action ($\y^r = 0$) region of the video. 

\vspace{3pt}\noindent\textbf{Local and global feature encoding.}
Let $E$ be the video encoder that transforms a clip $\X$ into a feature vector $f$ of size $F$. We refer to $f$ as the local clip feature. 
Let $\{\X_i\}$ be the set of clips from an untrimmed video. We refer to the max-pooled feature $f^{g} = \max(E(\X_i))$ as the global video feature (GVF). 
Given only a short clip $\X$, it is challenging to classify whether $\X$ is inside or outside an action. The challenge stems from the fact that we only have access to local context, while the task we wish to solve inherently requires global understanding of the video content. To overcome this challenge, we combine the GVF with the local clip feature to better learn the task. We can think of the GVF as a conditioning vector for deciding foreground \vs background. 
We study other GVF pooling functions in the appendix.

\vspace{3pt}\noindent\textbf{Two classification heads.}
We employ two classification heads to pretrain the encoder. Specifically, the first head (action label head) consists of a fully-connected (FC) layer $\W^c$ of size $F \times C$, where $C$ is the number of action classes in the dataset. $\W^c$ transforms the local features $f$ to an action label logits vector $\hat{\y}^c$. The second head (temporal region head) is an FC layer $\W^r$ of size $2F$$\times$$2$, which takes as input the concatenation of the local and global features, $f \oplus f^{g}$, to produce a temporal region logits vector $\hat{\y}^r$.

\vspace{3pt}\noindent\textbf{Loss.} 
We optimize our loss for each input clip $\X$:
\begin{align}
    \text{loss}= 
    \begin{cases}
        \alpha^r \mathcal{L}(\hat{\y}^r, \y^r) + \alpha^c \mathcal{L}(\hat{\y}^c, \y^c) , & \text{if } \y^r = 1\\
        \alpha^r \mathcal{L}(\hat{\y}^r, \y^r),              & \text{otherwise},
    \end{cases}
    \label{eq:loss}
\end{align}
\noindent where $\mathcal{L}$ is the cross-entropy loss and $(\alpha^c, \alpha^r)$ are trade-off coefficients to weigh the losses of the two heads. The loss is the sum of the two head losses when the clip is from the foreground, \ie $\y^r = 1$, and is the loss from the second head when the clip is from the background. 

\vspace{3pt}\noindent\textbf{Optimization details.}  
Temporally annotated video datasets have a natural imbalance between the temporal duration of foreground \vs background. To mitigate this imbalance, we subsample clips from videos in such a way that we train on the same number of foreground and background clip samples. 
We initialize our encoder weights with those pretrained on Kinetics-400~\cite{dataset_kinetics}. Many of the recent video architectures have publicly released their Kinetics-pretrained weights, and we make use of these models in our experiments. Ideally, we wish to backpropagate the loss (Equation~\ref{eq:loss}) through the GVF portion of our model. However, and as mentioned earlier, it is impractical to treat entire \emph{untrimmed} videos in commodity GPUs. Thus, we freeze the GVF during training, \ie we precompute the GVF of each video from the Kinetics-pretrained initialized encoder.

\section{Experiments}\label{sec:experiments}

\begin{table*}[t!]
    \small
    \centering
    \tabcolsep=0.2cm
    \caption{\textbf{Effects of TSP on target tasks.} We compare features pretrained with our TSP task \vs those pretrained with \textit{TAC on Kinetics} and \textit{TAC on ActivityNet}. We use R(2+1)D-34 encoders and pretrain on ActivityNet. We use G-TAD~\cite{xu2020gtad}, BMN~\cite{Lin_2019_ICCV}, and BMT~\cite{bmt} as algorithms for the ActivityNet TAL, Proposals, and Dense-Captioning tasks, respectively. The column corresponding to the main evaluation metric for each task is highlighted in grey and the best performance is in bold. TSP significantly outperforms the baselines on all tasks.}
    \vspace{-7pt}
    \begin{tabular}{l|cccg|cccg|ccg}
\toprule
     Video Task           & \multicolumn{4}{c|}{Temporal Action Localization } & \multicolumn{4}{c|}{Action Proposal Generation} & \multicolumn{3}{c}{Dense Video Captioning}      \\
    Feature Pretraining  &\footnotesize 0.5   &\footnotesize 0.75  &\footnotesize 0.95 & Avg. &\footnotesize AR@1  &\footnotesize AR@10 &\footnotesize AR@100 & AUC &\footnotesize BLEU@3 &\footnotesize BLEU@4 & METEOR \\\midrule
    TAC on Kinetics      &   48.54 &   34.24 &   7.85 &   33.32  &   34.19 &   57.52 &   75.56 &   67.91 &   3.42 &   1.58 &   8.17 \\
    TAC on ActivityNet   &   49.76 &   34.87 &   8.65 &   34.08  &   34.67 &   57.89 &   75.65 &   68.08 &   3.63 &   1.74 &   8.21 \\
    TSP w/o GVF          &\bf51.45 &   36.87 &   9.11 &   35.75  &   34.97 &\bf59.35 &   76.47 &   68.88 &   3.75 &   1.83 &   8.42 \\
    TSP on ActivityNet   &   51.26 &\bf37.12 &\bf9.29 &\bf35.81  &\bf34.99 &   58.96 &\bf76.63 &\bf69.04 &\bf4.16 &\bf2.02 &\bf8.75 \\
\bottomrule
    \end{tabular}
    \vspace{-3pt}
    \label{table:main_results}
\end{table*}

\subsection{Experimental Settings}

\noindent\textbf{Pretraining datasets.}
To pretrain with our TSP strategy, we need a dataset of untrimmed videos with temporal boundary annotations. Thus, we leverage two standard datasets: ActivityNet v1.3~\cite{dataset_activitynet} and THUMOS14~\cite{dataset_thumos14}.
\textbf{\textit{ActivityNet}}: This dataset has $20$K untrimmed videos and $200$ activity classes. It is split into training, validation, and testing subsets, where the testing subset labels are withheld for an annual challenge. Following standard practices, we use the training subset ($10024$ videos) to train and the validation subset ($4926$ videos) to test.
\textbf{\textit{THUMOS14}}: This dataset has $1010$ validation and $1574$ testing videos annotated with $101$ sport-related action classes at the video-level. Among these videos, only $200$ validation and $213$ testing videos have temporal annotations for $20$ sport actions. We use these $200$ validation videos to train and the $213$ testing videos to test.

\vspace{3pt}\noindent\textbf{Encoder architectures.}
We conduct experiments using two architectures: ResNet3D and R(2+1)D~\cite{tran2018closer}. We select these backbones for their recognized good performance, speed, and efficiency.
\textbf{\textit{ResNet3D}}: This is the 3D version of the 2D ResNet~\cite{he2016deep} CNN for images. ResNet3D is composed of a series of 3D convolution layers with residual skip connections. In our experiments and for simplicity, we consider the $18$-layer variant of ResNet3D.
\textbf{\textit{R(2+1)D}}: This encoder is also a ResNet-based backbone. It decomposes each spatio-temporal 3D convolution kernel into a 2D (spatial) and a 1D (temporal) convolution. Compared to ResNet3D, R(2+1)D is more efficient and light-weight, and it has been shown to maintain high performance on video tasks. In our experiments, we use the $18$ and $34$-layer versions of R(2+1)D.

\vspace{3pt}\noindent\textbf{Implementation details.}
In order to cope with the diversity of video formats present in ActivityNet and THUMOS14, we re-encode all videos in MP4 format with a constant frame rate of $30$ fps. We sample clips of $L=16$ frames with a stride of $2$ frames, such that each clip covers a temporal receptive field of approximately one second. While keeping the aspect ratio fixed, frames are resized such that the smallest dimension is $128$ pixels and then cropped to $H \times W = 112 \times 112$ pixels, randomly in training but deterministically centered during testing.
The videos are split into temporally contiguous segments, representing foreground (action) and background (no action) content. We select $5$ clips per segment, sampled randomly (temporal jittering) during training and uniformly in testing.
We set $\alpha^c=\alpha^r=1$ in Equation~(\ref{eq:loss}), and use a distributed SGD optimizer with different learning rates per module: $10^{-4}$ for the video encoder and a grid search among [$0.002$, $0.004$, $0.006$, $0.008$, $0.01$] for the two classification heads. We train for $8$ epochs with a batch size of $32$ clips per GPU. We use two V100 GPUs and scale the learning rate linearly with the number of GPUs. We use a linear learning rate warm up strategy over the first $2$ epochs and decay factor of $\gamma = 0.01$ at epochs $4$ and $6$. We select the best model among learning rates and training epochs based on the average validation clip accuracy of the two classification heads.

\vspace{3pt}\noindent\textbf{Baselines.}
We compare our pretraining approach with TAC pretraining. In particular, we consider the following baselines: \textit{TAC on Kinetics}, \textit{TAC on ActivityNet}, and \textit{TAC on THUMOS14}. The models from the second and third baselines are finetuned from a Kinetics-pretrained model.

\vspace{3pt}\noindent\textbf{Target tasks and evaluation metrics.}
We consider three localization tasks to evaluate TSP pretrained features: TAL on both ActivityNet and THUMOS14, Proposals on ActivityNet, and Dense-Captioning on ActivityNet Captions~\cite{activitynet_captions_dataset}. 
For the TAL tasks, the performance is measured using the mean Average Precision (mAP) metric, where a predicted temporal segment is considered a true positive, if it satisfies a temporal Intersection over Union (tIoU) threshold with a ground truth instance of the correct action label. Following standard practice, we use the average mAP over tIoUs $[0.5:0.05:0.95]$ as the main metric for ActivityNet and the mAP at tIoU=$0.5$ (mAP@$0.5$) for THUMOS14. 
For the Proposals task, the main evaluation metric is the area under the curve (AUC) of the average recall (AR) \vs average number of proposals per video. Following common practice in ActivityNet, we limit the number of proposals to $100$ per video when computing the AUC. We also report AR at $1$, $10$, and $100$ proposals as additional metrics.
Following common practice in the Dense-Captioning task, we use BLEU@$3$, BLEU@$4$, and METEOR averaged over tIoUs $[0.3, 0.5, 0.7, 0.9]$ to evaluate performance.

\vspace{3pt}\noindent\textbf{Algorithms for the target tasks.}
In order to showcase the benefits of TSP pretrained features compared to the baselines, we retrain a variety of state-of-the-art algorithms for each target task atop features extracted from TSP pretrained encoders as well as the baseline encoders. We select the algorithms based on (1) their strong performance on the target tasks and (2) the availability of open-sourced code. Here, we briefly discuss each algorithm and how we apply it to our features. \emph{It is essential to note that we do not innovate in any of these algorithms, and we use their default hyperparameter settings unless otherwise stated below}. We simply swap the visual features they originally use with ours or those of the encoder baselines we compare against. 
\textbf{\textit{G-TAD}}~\cite{xu2020gtad}: 
We use G-TAD for TAL on both ActivityNet and THUMOS14. G-TAD originally uses a Kinetics-pretrained TSN~\cite{tsn} encoder to extract RGB and Flow features, then trains on their concatenation. For G-TAD on THUMOS14, we increase the very small default learning rate by $\times10$ (\ie to $0.0004$) to speed up the training.
\textbf{\textit{BMN}}~\cite{Lin_2019_ICCV}:
BMN is used for both Proposals and TAL on ActivityNet. BMN did not release code for THUMOS14, and it uses the same precomputed features as G-TAD.
\textbf{\textit{P-GCN}}~\cite{Zeng_2019_ICCV}:
We employ P-GCN for TAL on THUMOS14. P-GCN did not release code for ActivityNet. P-GCN extracts features from an RGB and Flow I3D Kinetics-pretrained encoder. Then, two RGB and Flow localization models are trained \textit{independently} and their results are combined at inference time. We keep the Flow model unchanged and only retrain the RGB model with our features.
\textbf{\textit{BMT}}~\cite{bmt}:
BMT is used for the Dense-Captioning task on the ActivityNet Captions dataset. BMT uses visual and audio features. The visual features are the summation of RGB and Flow features from I3D Kinetics-pretrained encoders, and the audio features are from a VGG-like encoder pretrained on AudioSet~\cite{dataset_audioset}. We keep the audio features as is and replace the visual features with ours.

\subsection{Ablation Study}
Here, we extensively ablate TSP along four dimensions: target localization task, encoder architecture, localization algorithm, and pretraining dataset.

\vspace{3pt}\noindent\textbf{Study 1: Effects of TSP on target tasks.}
This study aims to compare features pretrained with TSP \vs those pretrained with the baselines, \textit{TAC on Kinetics} and \textit{TAC on ActivityNet}, on multiple target tasks. Specifically, we pretrain with the ActivityNet dataset and use an R(2+1)D-34 for the baseline encoders as well as our own. We use G-TAD, BMN, and BMT as the algorithms for the ActivityNet TAL, Proposals, and Dense-Captioning tasks, respectively. Table~\ref{table:main_results} summarizes the results. 
\textit{\textbf{Observations:}}\begin{table}[t!]
    \small
    \centering
    \caption{\textbf{Contribution of each TSP classification head to the target task performance.} We pretrain R(2+1)D-34 on ActivityNet and test the features on ActivityNet TAL using G-TAD~\cite{xu2020gtad}.}
    \vspace{-7pt}
    \begin{tabular}{l|cccg}
\toprule
Feature Pretraining     &      0.5 &     0.75 &    0.95 &    Avg \\\midrule
TSP w/o Temporal Region &    49.76 &    34.87 &    8.65 &    34.08  \\
TSP w/o Action Label    &    51.23 &    36.79 &\bf 9.91 &    35.72 \\ 
TSP                     &\bf 51.26 &\bf 37.12 &    9.29 &\bf 35.81  \\
\bottomrule
\end{tabular}
\vspace{-8pt}
\label{table:ablation_TSP_heads}
\end{table}%
\textbf{(I)} \textit{TAC on ActivityNet} outperforms \textit{TAC on Kinetics} for all three tasks. This makes sense given the fact that the former baseline is pretrained on the same dataset used in the target tasks. However, TSP features consistently show the best performance across all tasks. Specifically, TSP outperforms both baselines by at least +$1.73$\% in average mAP on TAL, +$0.96$\% in AUC on Proposals, and +$0.54$\% in average METEOR on Dense-Captioning. These significant gains underscore the effectiveness of TSP pretraining in encoding better temporal representations for untrimmed videos. 
\textbf{(II)} On the TAL task, TSP features significantly boost performance at high tIoU thresholds (\eg mAP@$0.75$ is $37.12$\% for TSP \vs $34.87$\% for \textit{TAC on ActivityNet}). Better mAP at high tIoUs signifies tighter temporal predictions around the ground truth action instances. This indicates that TSP pretrained features can encode better boundary contrast between the action and its nearby background context.
\textbf{(III)} While TSP pretraining without the GVF (\textit{TSP w/o GVF} in the table) outperforms the baselines, using GVF for the second classification head consistently boosts performance across all tasks (\eg $8.75$\% \vs $8.42$\% in average METEOR on Dense-Captioning). This validates our design choice and shows the importance of GVF in helping the local features be more temporally-sensitive. Given this observation, we omit \textit{TSP w/o GVF} from the remaining ablation studies.\begin{table*}[ht!]
    \small
    \centering
    \caption{\textbf{TSP for different video encoders.} We pretrain ResNet3D-18, R(2+1)D-18, and R(2+1)D-34 on ActivityNet and compare the features on the ActivityNet TAL task using G-TAD~\cite{xu2020gtad} as the TAL algorithm. Our TSP features consistently outperform the baselines for every encoder type, indicating the generalizability of our pretraining to different backbone architectures.} 
    \vspace{-6pt}
    \begin{tabular}{l|cccg|cccg|cccg}
\toprule
 Backbone Architecture & \multicolumn{4}{c|}{ ResNet3D-18}   & \multicolumn{4}{c|}{ R(2+1)D-18}    & \multicolumn{4}{c}{ R(2+1)D-34}    \\
Feature Pretraining    & 0.5   & 0.75  & 0.95 & Avg.  & 0.5   & 0.75  & 0.95 & Avg.  & 0.5   & 0.75  & 0.95 & Avg.  \\\midrule
TAC on Kinetics        &   47.97 &   33.21 &\bf8.96 &   32.78  &   47.57 &   33.11 &   8.10 &   32.46  &   48.54 &   34.24 &   7.85 &   33.32  \\
TAC on ActivityNet     &   48.71 &   34.22 &   8.82 &   33.40  &   49.00 &   34.56 &\bf9.42 &   33.87  &   49.76 &   34.87 &   8.65 &   34.08  \\
TSP on ActivityNet &\bf49.81 &\bf34.81 &   8.63 &\bf34.10  &\bf50.07 &\bf35.61 &   8.96 &\bf34.71  &\bf51.26 &\bf37.12 &\bf9.29 &\bf35.81  \\
\bottomrule
    \end{tabular}
    \label{table:ablation_backbone_architecture}
\end{table*}%
\textbf{(IV)} While the Dense-Captioning experiment is conducted on the same pretraining videos, the ActivityNet Captions temporal annotations~\cite{activitynet_captions_dataset} used for training the Dense-Captioning methods do not necessarily align with the ActivityNet temporal action annotations used for our pretraining. Nevertheless, TSP still provides an improvement over the baselines.
\textbf{(V)} Table~\ref{table:ablation_TSP_heads} studies the contribution of each TSP classification head to the target task performance. We observe that the performance boost comes mostly from the temporal region head, validating the importance of pretraining on foreground and background clips to attain temporal-sensitivity.

\vspace{3pt}\noindent\textbf{Study 2: TSP for different video encoders.}
This experiment explores TSP pretraining for different video architectures. Specifically, we pretrain ResNet3D-18, R(2+1)D-18, and R(2+1)D-34 on ActivityNet and compare the features on the ActivityNet TAL task using G-TAD as the TAL algorithm (refer to Table~\ref{table:ablation_backbone_architecture}).
\textit{\textbf{Observations:}}
\textbf{(I)} We observe similar performance trends among the different pretraining strategies regardless of the encoder type. In particular, our TSP features successfully outperform the baselines for every encoder. This indicates the generalization capability of the TSP pretraining to different backbones. 
\textbf{(II)} Aligned with observations made by previous works~\cite{tran2018closer}, R(2+1)D-18 exhibits better performance compared to ResNet3D-18 (average mAP of $34.71$\% \vs $34.10$\%). 
\textbf{(III)} Not only does the deeper R(2+1)D-34 pretrained with the TSP strategy achieve better performance compared to R(2+1)D-18, but interestingly, the performance gap between TSP and \textit{TAC on Kinetics} widens with the deeper encoder (+$2.25$\% for R(2+1)D-18 \vs +$2.49$\% for R(2+1)D-34). Similarly, TSP performance gap with \textit{TAC on ActivityNet} increases from +$0.84$\% for R(2+1)D-18 to +$1.73$\% for R(2+1)D-34. This suggests that our pretraining can potentially show even larger gains for more sophisticated and deeper encoders.

\vspace{3pt}\noindent\textbf{Study 3: TSP with other localization algorithms.}
We investigate here whether TSP features can consistently improve performance on the target task, regardless of the localization algorithm used. To that end, we conduct the same TAL on ActivityNet experiment from Study 1 (cf. Table~\ref{table:main_results}) but with the BMN algorithm instead of G-TAD. Table~\ref{table:ablation_tal_anet_bmn} summarizes the results using BMN.
\textit{\textbf{Observations:}}
\textbf{(I)} Our TSP features used with BMN show similar performance gains as when they are used with G-TAD, with at least a $0.92$\% gap in average mAP with the TAC-based pretrainings. This demonstrates that our features are more discriminative for the task and that they can benefit different algorithms.
\textbf{(II)} Both BMN and G-TAD originally use the same features (TSN pretrained on Kinetics) and have a $0.24$\% gap in average mAP. However, when both are trained using TSP features, BMN bridges the performance gap with G-TAD to be only $0.14$\%. This highlights the importance of having temporally-sensitive video features for localization tasks.

\begin{table}[t!]
    \small
    \centering
    \caption{\textbf{TSP with other localization algorithms.} We conduct the same TAL on ActivityNet experiment from Table~\ref{table:main_results} but with the BMN algorithm instead of G-TAD. Our TSP features achieve the best performance when used with BMN as well.
    }
    \vspace{-6pt}
    \begin{tabular}{l | c c c g }
        \toprule
        Feature Pretraining &    0.5  &   0.75  &   0.95 &   Avg.  \\\midrule
        TAC on Kinetics     &   49.95 &   35.31 &   8.61 &   34.46 \\
        TAC on ActivityNet  &   50.78 &   35.40 &   7.96 &   34.75 \\
        TSP on ActivityNet  &\bf51.23 &\bf36.78 &\bf9.50 &\bf35.67 \\
        \bottomrule
    \end{tabular}
    \label{table:ablation_tal_anet_bmn}
\end{table}

\begin{table}[t!]
    \small
    \centering
    \caption{\textbf{TSP on different datasets.} We pretrain R(2+1)D-34 on THUMOS14 and on ActivityNet, and use P-GCN~\cite{Zeng_2019_ICCV} and G-TAD~\cite{xu2020gtad} for the TAL task on THUMOS14. TSP features are applicable to and transferable across different datasets.}
    \label{table:ablation_tal_thumos14}
    \vspace{-6pt}
    \begin{subtable}{\linewidth}
        \centering
        \tabcolsep=0.19cm
    	\caption{\textbf{P-GCN}. Results are reported for the RGB model / RGB+Flow models.}
    	\vspace{-5pt}
        \begin{tabular}{l | c g c}
            \toprule
            Feature Pretraining & 0.3         & 0.5      & 0.7  \\\midrule
            TAC on Kinetics     & 52.4 / 65.9 & 37.8 / 49.0 & 15.6 / 22.9 \\
            TSP on ActivityNet  & 54.2 / 65.4 & 39.4 / 51.0 & 14.7 / 22.2 \\\midrule
            TAC on THUMOS14     & 54.4 / 66.4 & 38.7 / 50.0 & 16.1 / 23.3 \\
            TSP on THUMOS14     &\bf 58.0 / 69.1 &\bf 44.2 / 53.5 &\bf 18.5 / 26.0 \\
            \bottomrule
        \end{tabular}
    \label{table:ablation_tal_thumos14_pgcn}
    \end{subtable}

    \begin{subtable}{\linewidth}
        \centering
        \vspace{3pt}
    	\caption{\textbf{G-TAD}}
    	\vspace{-5pt}
        \begin{tabular}{l | c c g c c}
            \toprule
            Feature Pretraining &   0.3  &   0.4  &   0.5  &   0.6  &   0.7  \\\midrule
            TAC on Kinetics     &   50.6 &   43.2 &   34.5 &   24.1 &   15.5 \\
            TSP on ActivityNet  &   53.4 &   45.9 &   37.0 &   26.7 &   16.1 \\\midrule
            TAC on THUMOS14     &   52.6 &   45.5 &   35.8 &   26.2 &   15.6 \\
            TSP on THUMOS14     &\bf59.6 &\bf52.0 &\bf43.2 &\bf32.2 &\bf21.1 \\
            \bottomrule
        \end{tabular}
        \label{table:ablation_tal_thumos14_gtad}
    \end{subtable}
\end{table}

\begin{table*}[t!]
    \small
    \centering
    \tabcolsep=0.13cm
    \caption{\textbf{SOTA comparison for TAL and Dense-Captioning.} We compare TSP with SOTA methods for (a) TAL on ActivityNet, (b) TAL on THUMOS14, and (c) Dense-Captioning on ActivityNet Captions. We use G-TAD~\cite{xu2020gtad}, P-GCN~\cite{Zeng_2019_ICCV}, and BMT~\cite{bmt} as the algorithms trained atop our features for each task, respectively. TSP achieves SOTA performance on (a) and (b) and is competitive on (c).}
    \vspace{-6pt}
\begin{subtable}{0.33\linewidth}
    \centering
    \caption{TAL on ActivityNet}
    \vspace{-5pt}
    \begin{tabular}{l|cccg}
    \toprule
Method                      & 0.5   & 0.75  & 0.95  & Avg.   \\ \midrule
C-TCN~\cite{li2020deep}      &   47.60 &   31.90 &   6.20  &   31.10 \\
P-GCN~\cite{Zeng_2019_ICCV}  &   48.26 &   33.16 &   3.27  &   31.11 \\
BMN~\cite{Lin_2019_ICCV}     &   50.07 &   34.78 &   8.29  &   33.85 \\
GTAN~\cite{long2019gaussian} &   52.61 &   34.14 &   8.91  &   34.31 \\
PBRNet~\cite{liu2020progressive} &\bf53.96 &   34.97 &   8.98  &   35.01 \\
\midrule
G-TAD~\cite{xu2020gtad}      &   50.36 &   34.60 &   9.02  &   34.09 \\
\bf TSP (ours)               &   51.26 &\bf37.12 &\bf9.29  &\bf35.81 \\
\bottomrule
    \end{tabular}
    \label{table:sota_tal_anet}
\end{subtable}
~~~
\begin{subtable}{0.36\linewidth}
    \centering
    \caption{TAL on THUMOS14}
    \vspace{-5pt}
    \begin{tabular}{l|ccgcc}
    \toprule
Method                       &   0.3  &   0.4  &   0.5  &   0.6  &   0.7  \\
\midrule
G-TAD~\cite{xu2020gtad}      &   54.5 &   47.6 &   40.2 &   30.8 &   23.4 \\ 
TAL-Net~\cite{chao_cvpr_2018}&   53.2 &   48.5 &   42.8 &   33.8 &   20.8 \\ 
Zhao~\etal~\cite{zhao2020bottom}&53.9 &   50.7 &   45.4 &   38.0 &   28.5 \\ 
PBRNet~\cite{liu2020progressive} &   58.5 &   54.6 &   51.3 &   41.8 &\bf29.5 \\
TSA-Net~\cite{TSA_Net}       &   65.6 &   61.4 &   53.0 &\bf42.4 &   28.8 \\
\midrule
P-GCN~\cite{Zeng_2019_ICCV}  &   63.6 &   57.8 &   49.1 &   --   &   --   \\ 
\bf TSP  (ours)              &\bf69.1 &\bf63.3 &\bf53.5 &   40.4 &   26.0 \\
\bottomrule
    \end{tabular}
    \label{table:sota_tal_thumos}
\end{subtable}
~~~
\begin{subtable}{0.26\linewidth}
    \centering
    \caption{Dense-Captioning}
    \vspace{-5pt}
    \begin{tabular}{l|ccg}
\toprule
Method              &   B@3  &   B@4  & M   \\ \midrule
Bi-SST~\cite{Wang_2018_CVPR}      &   2.27 &   1.13 &   6.10 \\ 
DVC~\cite{Li_dvc}   &   2.27 &   0.73 &   6.93 \\ 
MFT~\cite{mft}      &   2.82 &   1.24 &   7.08 \\ 
MDVC~\cite{mdvc}    &   2.60 &   1.07 &   7.31 \\ 
SDVC~\cite{sdvc}    &   2.94 &   0.93 &\bf8.82 \\
\midrule
BMT~\cite{bmt}      &   3.84 &   1.88 &   8.44 \\
\bf TSP (ours)      &\bf4.16 &\bf2.02 &   8.75 \\
\bottomrule 
    \end{tabular}
    \label{table:sota_captioning}
\end{subtable}
\label{table:sota}
\end{table*}

\begin{table}[t!]
    \small
    \centering
    \tabcolsep=0.1cm
    \caption{\textbf{SOTA comparison for Proposals on ActivityNet}. We use BMN atop our features. TSP significantly improves over BMN original performance and is competitive with SOTA.}
    \vspace{-6pt}
    \begin{tabular}{l|ccccc|cc}
    \toprule
 Method  & \cite{mgg}   & \cite{zhao2020bottom} & \cite{bcgcn} & \cite{dbg} & \cite{gao2020accurate} & BMN~\cite{Lin_2019_ICCV}  &\bf TSP \\\midrule
 \footnotesize AR@100  & 74.54            & 75.27                            & 76.73              & 76.65          &\bf78.63                       & 75.01          &   76.63 \\
\rowcolor{Gray}
 AUC & 66.43            & 66.51                            & 68.05              & 68.23          &\bf69.93                       & 67.10          &   69.04 \\
    \bottomrule
    \end{tabular}
    \label{table:sota_proposals}
\end{table}

\vspace{3pt}\noindent\textbf{Study 4: TSP on different datasets.}
Here, we study two aspects of TSP: its applicability to other pretraining datasets (\ie TSP pretrained on THUMOS14 and tested for TAL on THUMOS14), and its transferability across datasets (\ie TSP pretrained on ActivityNet and tested for TAL on THUMOS14).
Specifically, we pretrain R(2+1)D-34 on THUMOS14 and on ActivityNet, then apply P-GCN and G-TAD atop TSP features for the TAL task on THUMOS14. Table~\ref{table:ablation_tal_thumos14} compares the two TSP features with the baselines, \textit{TAC on Kinetics} and \textit{TAC on THUMOS14}.
\textit{\textbf{Observations:}}
\textbf{(I)} THUMOS14 is different from ActivityNet in two key aspects: THUMOS14 is much smaller, and it has a higher background to foreground ratio (\ie actions are sparser in THUMOS14). Despite these differences, TSP on THUMOS14 improves over the TAC-based baselines by significant margins, regardless of the localization algorithm. Specifically when using P-GCN, TSP on THUMOS14 features improve the RGB model results by at least $5.5$\% in mAP@$0.5$. Moreover, combining the predictions of our newly-trained RGB model with that of the original (unchanged) Flow modality boosts the overall performance by at least $3.5$\% in mAP@$0.5$.
\textbf{(II)} Using TSP features pretrained on ActivityNet (TSP on ActivityNet) outperforms both \textit{TAC on Kinetics} and \textit{TAC on THUMOS14} in mAP@$0.5$. This shows that TSP features are transferable across TAL datasets.

\subsection{State-of-the-Art (SOTA) Comparison}
While the previous ablations shed light on the generalization of TSP across multiple tasks, video encoders, algorithms, and datasets, this subsection puts our results in perspective and compares them with SOTA algorithms for each localization task. We report the comparative results in Tables~\ref{table:sota} and~\ref{table:sota_proposals}. Note that we build TSP upon the best-performing \textit{publicly available} code for each task, namely G-TAD~\cite{xu2020gtad}, P-GCN~\cite{Zeng_2019_ICCV}, BMN~\cite{Lin_2019_ICCV}, and BMT~\cite{bmt}. 
In \textbf{\textit{TAL on ActivityNet}} (Table~\ref{table:sota_tal_anet}), we reach SOTA performance with TSP. We achieve $35.81$\% in average mAP, a boost of $0.80$\% \wrt the previous SOTA PBRNet~\cite{liu2020progressive} and a boost of $1.72$\% \wrt our baseline G-TAD~\cite{xu2020gtad}. Moreover, TSP (with RGB features only) outperforms SOTA methods~\cite{li2020deep,Lin_2019_ICCV,liu2020progressive,xu2020gtad,Zeng_2019_ICCV} that use RGB and Flow features.
In \textbf{\textit{TAL on THUMOS14}} (Table~\ref{table:sota_tal_thumos}), we achieve $53.5$\% in mAP@$0.5$, a boost of $0.5$\% \wrt the previous SOTA TSA-Net~\cite{TSA_Net} and a boost of $4.4$\% \wrt our baseline P-GCN~\cite{Zeng_2019_ICCV}. The results display different improvements on both datasets, focusing on higher tIoU for ActivityNet and lower tIoU on THUMOS14. We argue that this discrepancy originates from the different activity densities in both datasets.
In \textbf{\textit{Action Proposal Generation}} (Table~\ref{table:sota_proposals}), we reach $69.04$\% in AUC, a boost of $1.96$\% \wrt our baseline BMN~\cite{Lin_2019_ICCV}, but fall short of RapNet~\cite{gao2020accurate} ($-0.89$\%). 
In \textbf{\textit{Dense Video Captioning}} (Table~\ref{table:sota_captioning}), we reached $8.75$\% in average METEOR, a $0.31$\% improvement over the baseline BMT~\cite{bmt}, but fall short of SDVC~\cite{sdvc} ($-0.07$\%). We argue that SDVC~\cite{sdvc} uses a reinforcement learning paradigm that optimizes for the METEOR metrics directly, trading off BLEU performances to overfit on METEOR. In contrast, the TSP-empowered BMT model achieves balanced performances in both BLEU and METEOR metrics.

\begin{table}[t!]
    \small
    \centering
    \tabcolsep=0.20cm
    \caption{\textbf{SOTA SSL comparison.} We compare TSP with XDC for TAL on THUMOS14. Both use R(2+1)D-18 and G-TAD.} 
    \label{table:comparison_with_xdc}
    \vspace{-6pt}
    \begin{tabular}{l | c c g c c}
        \toprule
        Feature Pretraining &    0.3  &    0.4  &    0.5  &    0.6  &    0.7  \\\midrule
        TAC on Kinetics     &    45.4 &    38.9 &    30.5 &    19.7 &    11.5 \\
        TAC on THUMOS14       &    48.0 &    41.6 &    33.3 &    23.7 &    14.6 \\
        XDC on IG-Kinetics~\cite{alwassel2020self}  &    51.5 &    44.9 &    37.2 &    28.7 &\bf 20.0 \\
        TSP on THUMOS14       &\bf 57.1 &\bf 50.2 &\bf 41.0 &\bf 30.4 &    19.7 \\
        \bottomrule
    \end{tabular}
\end{table}

\subsection{Comparison with Self-Supervised Encoders}

Recent self-supervised learning (SSL) methods have shown impressive performance on video tasks such as action classification~\cite{avts,misra2016shuffle,motion_statistics,clip_order}. Here, we compare TSP features with SOTA SSL features for temporal localization tasks. Specifically, we compare with XDC~\cite{alwassel2020self}, a recent SOTA SSL method that learns video and audio features via cross-modal deep clustering. Table~\ref{table:comparison_with_xdc} compares TSP and XDC features for TAL on THUMOS14 under the same settings: R(2+1)D-18 encoder and G-TAD algorithm. Although XDC impressively outperforms the supervised TAC baselines, it falls short of TSP performance by $2.8\%$ in mAP@0.5. While it is expected that SSL requires more video data for pretraining than supervised pretraining, it is worthwhile to point out that XDC pretrains on $65$M videos from IG-Kinetics~\cite{ghadiyaram2019large}, \ie $260$ times more videos than TSP.
\section{Feature Analysis}
\label{sec:analysis}

\noindent We further analyze TSP pretrained features on ActivityNet.

\begin{table}[t!]
    \small
    \centering
    \caption{\textbf{Performance as a function of action length}. We report the performance of TAL on ActivityNet for different action lengths. TSP performs significantly better on Extra Short (XS) and Short (S) actions. XS and S make up about $70$\% of all actions.
    }
    \vspace{-6pt}
    \begin{tabular}{l|ccccc}
\toprule
    Instance Length    &   XS   &   S    &   M    &   L    &   XL   \\\midrule
    \% of the Dataset  &   53.7 &   16.2 &   16.8 &   9.7  &   4.0  \\\midrule
    TAC on Kinetics    &   16.4 &   41.3 &   53.2 &\bf68.4 &   72.3 \\
    TAC on ActivityNet &   17.5 &   42.0 &   53.1 &   67.5 &\bf72.5 \\
    TSP on ActivityNet &\bf19.3 &\bf44.2 &\bf53.9 &   67.8 &   71.3 \\
\bottomrule
    \end{tabular}
    \label{table:analysis_detad}
\vspace{-1pt}
\end{table}

\vspace{3pt}\noindent\textbf{DETAD analysis.}
Following DETAD~\cite{alwassel_eccv_2018}, we analyze the TAL on ActivityNet performance (average mAP) for five different groups of activities based on their length (Table~\ref{table:analysis_detad}): Extra Short (XS: (0s, 30s]), Short (S: (30s, 60s]), Medium (M: (60s, 120s]), Long (L: (120s, 180s]), and Extra Long (XL: $>$ 180s). The extra short instances, XS, are known to be the most challenging to localize \cite{alwassel_eccv_2018}, and they represent more than half of the annotated instances ($53.7$\%). Their temporal extent is limited as is the information available to recognize the activity. Such instances might be hidden among a significant amount of background.
It is clear that TAC performs well in localizing long activities, in particular because they are predominant in their corresponding videos. Yet, TAC achieves the worst performance on the challenging shorter activities. We argue that their localization is more sensitive to the classification of each single clip, since TAC is unaware of what an activity does \textit{not} look like in its temporal surrounding.
In contrast, TSP features outperform the TAC ones for the short activity instances (XS and S). We believe our learned clip feature is more aware of background, and thus more perceptive of temporal activity boundaries for localization. 
As a trade-off, it appears that TSP does not perform as well on extra long activities. We believe those long activities might include intermediate clips with content leaning toward a background activity, thus misleading the localization and resulting in slightly worse performance. Nevertheless, the XL activities merely represent $4.0$\% of the dataset, so the overall impact on performance is insignificant.

\begin{figure}[t]
    \centering
    (a)~~~~~~~~~~~~~~~~~~
    (b)~~~~~~~~~~~~~~~~~~
    (c)~~~~~~~~~~~~~~~~~~
    (d)\\
    \includegraphics[width=0.2425\linewidth]{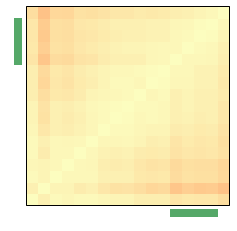}
    \includegraphics[width=0.2425\linewidth]{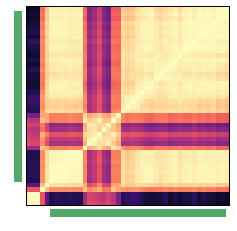}
    \includegraphics[width=0.2425\linewidth]{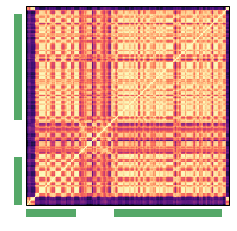}
    \includegraphics[width=0.2425\linewidth]{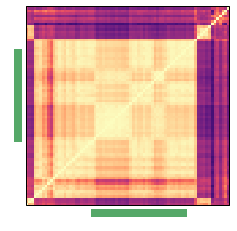} 
    \\
    \includegraphics[width=0.2425\linewidth]{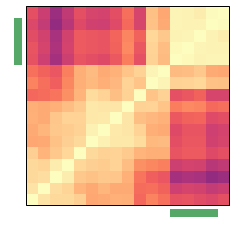}
    \includegraphics[width=0.2425\linewidth]{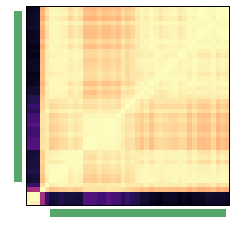}
    \includegraphics[width=0.2425\linewidth]{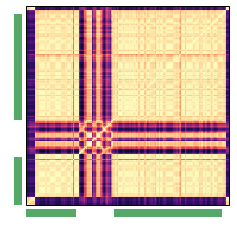}
    \includegraphics[width=0.2425\linewidth]{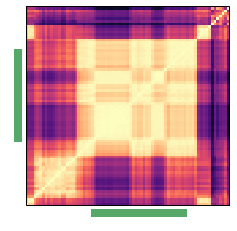}
    \\
    \includegraphics[width=0.66\linewidth]{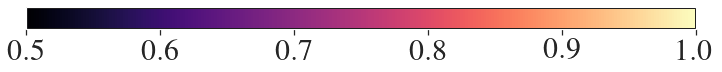}
    \vspace{-5pt}
    \caption{\textbf{Feature similarity}. Each column shows the similarity matrices for clips in a single video using \textit{TAC on Kinetics} (top) and TSP (bottom) features. The green lines represent the temporal extent of ground truth actions. Better viewed in color.}
    \vspace{-5pt}
    \label{fig:cosine_similarity}
\end{figure}

\vspace{3pt}\noindent\textbf{Feature similarity among video clips.}
Here, we analyze the similarity between video clip features within the same video. We expect the clip features from the same activity to be very similar (consensus), yet very different from the background clips in its temporal surrounding (sharpness).
Figure~\ref{fig:cosine_similarity} visualizes the cosine similarity between clips of the same video using \textit{TAC on Kinetics} \vs TSP features (more examples are in the \emph{supplementary material}). In (a), \textit{TAC on Kinetics} shows a high similarity between the activity and the background. This will inevitably make localization more difficult. In comparison, TSP better discriminates between background and activity. In (b), it appears that \textit{TAC on Kinetics} is trying to split the activity in two. By learning what background is and what it is not, TSP homogenizes the similarity between all clip features in the foreground activity. In (c), the TSP video encoder increases the differences between background and foreground features. It homogenizes the features within both activities (bottom left and top right corners), yet it does not enforce background features to be similar, resulting in an increase in dissimilarity within the background (see the apparent diagonal in the central square).  In (d), \textit{TAC on Kinetics} displays a high similarity for the clips of the foreground activity, yet they might look similar to the remaining background. TSP learns obvious dissimilarity between background and foreground. 
The TAC pretraining is unaware of the existence of background clips. As a result, it might recognize the class of some actions but is unable to localize them precisely. In contrast, TSP makes the encoder aware of the existence of background, and so the clip features across the video tend to be more informative for localization. Thus, TSP improves the encoder's discriminative ability, reduces the smoothing over the temporal axis, and leads to sharper localization.

\section{Conclusion}\label{sec:conclusion}
We present TSP, a novel temporally-sensitive supervised pretraining for video encoders, which not only trains to classify actions, but also considers background clips and global information to gain temporal sensitivity. We show that TSP features improve SOTA methods on the TAL, Proposals, and Dense-Captioning tasks. We argue TSP features can be preferred over other features to build more accurate models.

\noindent\textbf{Acknowledgments.}
This work is supported the King Abdullah University of Science and Technology (KAUST) Office of Sponsored Research (OSR) under Award No. OSR-CRG2017-3405.

{\small
\bibliographystyle{ieee_fullname}
\bibliography{egbib}
}

\onecolumn
\section*{Supplementary Material}
\appendix

\section{Pooling Function for GVF}
Table~\ref{table:supp_mat_max_vs_avg_GVF} compares between the performance of TSP with max-pooled \vs average-pooled GVF on the three target tasks: TAL, Proposals, and Dense-Captioning. TSP with max-pooled GVF offers better performance across all the tasks.

\begin{table*}[h!]
    \small
    \centering
    \caption{\textbf{Effects of GVF pooling function on target tasks.} We compare features pretrained with TSP using average-pooled \vs max-pooled GVF. We use R(2+1)D-34 encoders and pretrain on ActivityNet. We use G-TAD, BMN, and BMT as the methods for the ActivityNet TAL, Proposals, and Dense-Captioning tasks, respectively. TSP with max-pooled GVF is better on all tasks.}
    \begin{tabular}{l|cccg|cccg|ccg}
\toprule
 Video Task           & \multicolumn{4}{c|}{Temporal Action Localization } & \multicolumn{4}{c|}{Action Proposal Generation} & \multicolumn{3}{c}{Dense Video Captioning}      \\
Feature Pretraining  &\footnotesize 0.5   &\footnotesize 0.75  &\footnotesize 0.95 & Avg. &\footnotesize AR@1  &\footnotesize AR@10 &\footnotesize AR@100 & AUC &\footnotesize BLEU@3 &\footnotesize BLEU@4 & METEOR \\\midrule
TSP (avg GVF)        &   51.04 &   37.07 &\bf9.54 &   35.74  &   34.88 &\bf59.32 &   76.40 &   68.85 &   3.56 &   1.71 &   8.17\\
TSP (max GVF)        &\bf51.26 &\bf37.12 &   9.29 &\bf35.81  &\bf34.99 &   58.96 &\bf76.63 &\bf69.04 &\bf4.16 &\bf2.02 &\bf8.75 \\
\bottomrule
    \end{tabular}
    \label{table:supp_mat_max_vs_avg_GVF}
\end{table*}

\section{Extended Ablation Study Results}
In this section, we provide further statistical analysis for our studies on TSP. Statistical analyses are of great importance when comparing the performances of different algorithms. In particular, it helps us to understand whether a given improvement is significant or it is within the noise range. For each ablation study in the main paper, we reported the \textit{maximum} value over 5 runs, following common practice in the field. However, such practice might be misleading under certain scenarios, in particular if a proposed approach has high performance variance but has on average worse performance. To alleviate any doubts, and in an effort of transparency, we share here the \textit{mean} and \textit{standard deviation} performances for our proposed approach along with those of the TAC-pretrained baselines. Refer to Tables~\ref{table:supp_mat_extended_ablation_study_1}, \ref{table:supp_mat_extended_ablation_study_2}, \ref{table:supp_mat_extended_ablation_study_3}, and \ref{table:supp_mat_extended_ablation_study_4} for the extended statistical results of Study 1, 2, 3, and 4, respectively. These tables report the performance on \textit{all} tIoUs as well. The extended results show that our improvements are consistent with those reported in the main paper, and that such improvements do \textit{not} lie within the noise range. With such analysis, we can confidently say that TSP is \textit{statistically} better than the TAC-pretrained baselines.
\begin{table*}[ht]
    \small
    \centering
    \caption{\textbf{Effects of TSP on target tasks (extended results).} Each experiment in Study 1 is repeated five times, and we report the the mean, standard deviation (std), and max values over those five runs. Each table entry is given by \textit{\textbf{mean $\pm$ std (max)}}. The row/column corresponding to the main evaluation metric for each task is highlighted in grey and the best (\textit{mean}) performance is in bold.}
    \label{table:supp_mat_extended_ablation_study_1}
\begin{subtable}{\linewidth}
    \centering
    \caption{\bf TAL on ActivityNet using G-TAD with R(2+1)D-34.}
    \vspace{-5pt}
    \begin{tabular}{c|cccc}
    \toprule 
            & \multicolumn{4}{c}{Feature Pretraining} \\ 
    mAP@tIoU& TAC on Kinetics                           & TAC on ActivityNet                        & TSP w/o GVF                               & TSP on ActivityNet \\ \midrule
    0.50    & 48.269 {\scriptsize $\pm$ 0.241} (48.538) & 49.223 {\scriptsize $\pm$ 0.349} (49.761) & \bf 51.389 {\scriptsize $\pm$ 0.145} (51.445) & 51.206 {\scriptsize $\pm$ 0.162} (51.263) \\
    0.55    & 45.535 {\scriptsize $\pm$ 0.239} (45.900) & 46.588 {\scriptsize $\pm$ 0.346} (46.919) & \bf 48.532 {\scriptsize $\pm$ 0.107} (48.641) & 48.472 {\scriptsize $\pm$ 0.124} (48.551) \\
    0.60    & 42.824 {\scriptsize $\pm$ 0.284} (43.262) & 43.790 {\scriptsize $\pm$ 0.316} (44.131) & \bf 45.683 {\scriptsize $\pm$ 0.154} (45.872) & 45.637 {\scriptsize $\pm$ 0.109} (45.723) \\
    0.65    & 40.185 {\scriptsize $\pm$ 0.309} (40.617) & 41.111 {\scriptsize $\pm$ 0.297} (41.510) & 43.182 {\scriptsize $\pm$ 0.112} (43.212) & \bf 43.239 {\scriptsize $\pm$ 0.108} (43.327) \\
    0.70    & 37.487 {\scriptsize $\pm$ 0.285} (37.853) & 38.366 {\scriptsize $\pm$ 0.245} (38.631) & 40.451 {\scriptsize $\pm$ 0.135} (40.673) & \bf 40.534 {\scriptsize $\pm$ 0.169} (40.834) \\
    0.75    & 33.977 {\scriptsize $\pm$ 0.232} (34.241) & 34.780 {\scriptsize $\pm$ 0.154} (34.865) & 36.816 {\scriptsize $\pm$ 0.069} (36.865) & \bf 36.845 {\scriptsize $\pm$ 0.164} (37.123) \\
    0.80    & 30.191 {\scriptsize $\pm$ 0.158} (30.434) & 30.856 {\scriptsize $\pm$ 0.197} (30.874) & \bf 32.756 {\scriptsize $\pm$ 0.068} (32.865) & 32.678 {\scriptsize $\pm$ 0.118} (32.772) \\
    0.85    & 25.119 {\scriptsize $\pm$ 0.157} (25.299) & 25.820 {\scriptsize $\pm$ 0.155} (25.849) & 27.478 {\scriptsize $\pm$ 0.092} (27.620) & \bf 27.690 {\scriptsize $\pm$ 0.126} (27.712) \\
    0.90    & 19.152 {\scriptsize $\pm$ 0.164} (19.157) & 19.569 {\scriptsize $\pm$ 0.232} (19.617) & 20.998 {\scriptsize $\pm$ 0.095} (21.181) & \bf 21.375 {\scriptsize $\pm$ 0.152} (21.487) \\
    0.95    & 08.028 {\scriptsize $\pm$ 0.290} (07.847) & 08.274 {\scriptsize $\pm$ 0.429} (08.647) & 09.429 {\scriptsize $\pm$ 0.241} (09.109) & \bf 09.440 {\scriptsize $\pm$ 0.243} (09.286) \\ \midrule
\rowcolor{Gray}
    Average & 33.077 {\scriptsize $\pm$ 0.186} (33.315) & 33.837 {\scriptsize $\pm$ 0.189} (34.080) & 35.671 {\scriptsize $\pm$ 0.066} (35.748) &\bf 35.712 {\scriptsize $\pm$ 0.062} (35.808) \\
    \bottomrule
    \end{tabular}
\end{subtable}

\begin{subtable}{\linewidth}
    \centering
    \vspace{+5pt}
    \caption{\bf Proposals on ActivityNet using BMN with R(2+1)D-34.}
    \vspace{-5pt}
    \begin{tabular}{c|cccc}
    \toprule 
            & \multicolumn{4}{c}{Feature Pretraining} \\ 
    Metric  & TAC on Kinetics                           & TAC on ActivityNet                        & TSP w/o GVF                               & TSP on ActivityNet \\ \midrule
    AR@1    & 34.002 {\scriptsize $\pm$ 0.251} (34.185) & 34.452 {\scriptsize $\pm$ 0.152} (34.667) & 34.961 {\scriptsize $\pm$ 0.475} (34.971) &\bf 35.011 {\scriptsize $\pm$ 0.089} (34.991) \\
    AR@10   & 57.194 {\scriptsize $\pm$ 0.501} (57.520) & 57.772 {\scriptsize $\pm$ 0.149} (57.892) & 58.831 {\scriptsize $\pm$ 0.640} (59.346) &\bf 59.126 {\scriptsize $\pm$ 0.101} (58.961) \\
    AR@100  & 75.415 {\scriptsize $\pm$ 0.305} (75.561) & 75.654 {\scriptsize $\pm$ 0.067} (75.648) & 76.212 {\scriptsize $\pm$ 0.490} (76.469) &\bf 76.539 {\scriptsize $\pm$ 0.108} (76.627) \\ \midrule
\rowcolor{Gray}
    AUC     & 67.637 {\scriptsize $\pm$ 0.277} (67.912) & 67.959 {\scriptsize $\pm$ 0.087} (68.075) & 68.572 {\scriptsize $\pm$ 0.532} (68.875) &\bf 68.906 {\scriptsize $\pm$ 0.119} (69.035) \\
    \bottomrule
    \end{tabular}
\end{subtable}

\begin{subtable}{\linewidth}
    \centering
    \vspace{+5pt}
    \caption{\bf Dense-Captioning on ActivityNet Captions using BMT with R(2+1)D-34.}
    \vspace{-5pt}
    \begin{tabular}{l|ccg|ccg}
    \toprule 
                    & \multicolumn{3}{c|}{Ground Truth Proposals} & \multicolumn{3}{c}{Learned Proposals} \\ 
Feature Pretraining &\footnotesize BLEU@3 &\footnotesize BLEU@4 & METEOR &\footnotesize BLEU@3 &\footnotesize BLEU@4 & METEOR \\\midrule
TAC on Kinetics     &   4.32 &   1.76 &   10.93 &   3.42 &   1.58 &   8.17 \\
TAC on ActivityNet  &   4.64 &   1.94 &   10.99 &   3.63 &   1.74 &   8.21 \\
TSP w/o GVF         &\bf4.88 &\bf2.09 &   11.29 &   3.75 &   1.83 &   8.42 \\
TSP on ActivityNet  &   4.76 &   1.99 &\bf11.31 &\bf4.16 &\bf2.02 &\bf8.75 \\
    \bottomrule
    \end{tabular}
\end{subtable}

\end{table*}

\begin{table*}[t]
    \small
    \centering
    \caption{\textbf{TSP for different video encoders (extended results).} Each experiment in Study 2 is repeated five times, and we report the the mean, standard deviation (std), and max values over those five runs. Each table entry is given by \textit{\textbf{mean $\pm$ std (max)}}.}
    \label{table:supp_mat_extended_ablation_study_2}
\begin{subtable}{\linewidth}
    \centering
    \vspace{-5pt}
    \caption{\bf TAL on ActivityNet using G-TAD with ResNet3D-18.}
    \vspace{-5pt}
    \begin{tabular}{c|ccc}
    \toprule 
            & \multicolumn{3}{c}{Feature Pretraining} \\ 
    mAP@tIoU& TAC on Kinetics                           & TAC on ActivityNet                        & TSP on ActivityNet \\ \midrule
    0.50    & 47.514 {\scriptsize $\pm$ 0.310} (47.970) & 48.351 {\scriptsize $\pm$ 0.188} (48.708) &\bf 49.182 {\scriptsize $\pm$ 0.305} (49.806) \\
    0.55    & 44.629 {\scriptsize $\pm$ 0.323} (45.207) & 45.598 {\scriptsize $\pm$ 0.170} (45.926) &\bf 46.278 {\scriptsize $\pm$ 0.251} (46.683) \\
    0.60    & 42.005 {\scriptsize $\pm$ 0.377} (42.582) & 42.977 {\scriptsize $\pm$ 0.146} (43.178) &\bf 43.724 {\scriptsize $\pm$ 0.209} (44.069) \\
    0.65    & 39.312 {\scriptsize $\pm$ 0.359} (39.895) & 40.306 {\scriptsize $\pm$ 0.153} (40.339) &\bf 41.094 {\scriptsize $\pm$ 0.203} (41.374) \\
    0.70    & 36.515 {\scriptsize $\pm$ 0.329} (36.990) & 37.587 {\scriptsize $\pm$ 0.110} (37.627) &\bf 38.298 {\scriptsize $\pm$ 0.213} (38.541) \\
    0.75    & 32.878 {\scriptsize $\pm$ 0.253} (33.206) & 34.085 {\scriptsize $\pm$ 0.129} (34.217) &\bf 34.654 {\scriptsize $\pm$ 0.166} (34.814) \\
    0.80    & 29.014 {\scriptsize $\pm$ 0.207} (29.314) & 30.121 {\scriptsize $\pm$ 0.116} (30.205) &\bf 30.697 {\scriptsize $\pm$ 0.076} (30.702) \\
    0.85    & 24.459 {\scriptsize $\pm$ 0.111} (24.517) & 25.507 {\scriptsize $\pm$ 0.189} (25.602) &\bf 26.176 {\scriptsize $\pm$ 0.075} (26.182) \\
    0.90    & 18.808 {\scriptsize $\pm$ 0.229} (19.197) & 19.402 {\scriptsize $\pm$ 0.123} (19.427) &\bf 20.268 {\scriptsize $\pm$ 0.121} (20.156) \\
    0.95    & 08.306 {\scriptsize $\pm$ 0.516} (08.955) & 08.424 {\scriptsize $\pm$ 0.270} (08.816) &\bf 08.661 {\scriptsize $\pm$ 0.435} (08.625) \\ \midrule
\rowcolor{Gray}
    Average & 32.344 {\scriptsize $\pm$ 0.273} (32.783) & 33.235 {\scriptsize $\pm$ 0.089} (33.404) &\bf 33.903 {\scriptsize $\pm$ 0.147} (34.095) \\
    \bottomrule
    \end{tabular}
\end{subtable}

\begin{subtable}{\linewidth}
    \centering
    \vspace{5pt}
    \caption{\bf TAL on ActivityNet using G-TAD with R(2+1)D-18.}
    \vspace{-5pt}
    \begin{tabular}{c|ccc}
    \toprule 
            & \multicolumn{3}{c}{Feature Pretraining} \\ 
    mAP@tIoU& TAC on Kinetics                           & TAC on ActivityNet                        & TSP on ActivityNet \\ \midrule
    0.50    & 47.218 {\scriptsize $\pm$ 0.297} (47.573) & 48.701 {\scriptsize $\pm$ 0.170} (49.003) &\bf 49.883 {\scriptsize $\pm$ 0.187} (50.069) \\
    0.55    & 44.445 {\scriptsize $\pm$ 0.289} (44.827) & 45.846 {\scriptsize $\pm$ 0.164} (46.157) &\bf 47.079 {\scriptsize $\pm$ 0.203} (47.226) \\
    0.60    & 41.715 {\scriptsize $\pm$ 0.297} (42.055) & 43.275 {\scriptsize $\pm$ 0.133} (43.475) &\bf 44.483 {\scriptsize $\pm$ 0.179} (44.433) \\
    0.65    & 38.992 {\scriptsize $\pm$ 0.273} (39.367) & 40.816 {\scriptsize $\pm$ 0.141} (40.977) &\bf 41.983 {\scriptsize $\pm$ 0.164} (41.909) \\
    0.70    & 36.233 {\scriptsize $\pm$ 0.270} (36.653) & 38.037 {\scriptsize $\pm$ 0.126} (38.202) &\bf 39.245 {\scriptsize $\pm$ 0.101} (39.243) \\
    0.75    & 32.762 {\scriptsize $\pm$ 0.216} (33.113) & 34.273 {\scriptsize $\pm$ 0.205} (34.562) &\bf 35.568 {\scriptsize $\pm$ 0.091} (35.608) \\
    0.80    & 28.919 {\scriptsize $\pm$ 0.242} (29.301) & 30.609 {\scriptsize $\pm$ 0.224} (30.743) &\bf 31.595 {\scriptsize $\pm$ 0.191} (31.987) \\
    0.85    & 24.481 {\scriptsize $\pm$ 0.163} (24.745) & 25.837 {\scriptsize $\pm$ 0.157} (25.993) &\bf 26.677 {\scriptsize $\pm$ 0.146} (26.885) \\
    0.90    & 18.691 {\scriptsize $\pm$ 0.170} (18.839) & 19.920 {\scriptsize $\pm$ 0.198} (20.119) &\bf 20.464 {\scriptsize $\pm$ 0.167} (20.773) \\
    0.95    & 08.111 {\scriptsize $\pm$ 0.574} (08.099) & 08.971 {\scriptsize $\pm$ 0.447} (09.424) &\bf 09.072 {\scriptsize $\pm$ 0.434} (08.958) \\ \midrule
\rowcolor{Gray}
    Average & 32.157 {\scriptsize $\pm$ 0.220} (32.457) & 33.629 {\scriptsize $\pm$ 0.161} (33.865) &\bf 34.605 {\scriptsize $\pm$ 0.101} (34.709) \\ 
    \bottomrule
    \end{tabular}
\end{subtable}

\begin{subtable}{\linewidth}
    \centering
    \vspace{+5pt}
    \caption{\bf TAL on ActivityNet using G-TAD with R(2+1)D-34.}
    \vspace{-5pt}
    \begin{tabular}{c|ccc}
    \toprule 
            & \multicolumn{3}{c}{Feature Pretraining} \\ 
    mAP@tIoU& TAC on Kinetics                           & TAC on ActivityNet                        & TSP on ActivityNet \\ \midrule
    0.50    & 48.269 {\scriptsize $\pm$ 0.241} (48.538) & 49.223 {\scriptsize $\pm$ 0.349} (49.761) &\bf 51.206 {\scriptsize $\pm$ 0.162} (51.263) \\
    0.55    & 45.535 {\scriptsize $\pm$ 0.239} (45.900) & 46.588 {\scriptsize $\pm$ 0.346} (46.919) &\bf 48.472 {\scriptsize $\pm$ 0.124} (48.551) \\
    0.60    & 42.824 {\scriptsize $\pm$ 0.284} (43.262) & 43.790 {\scriptsize $\pm$ 0.316} (44.131) &\bf 45.637 {\scriptsize $\pm$ 0.109} (45.723) \\
    0.65    & 40.185 {\scriptsize $\pm$ 0.309} (40.617) & 41.111 {\scriptsize $\pm$ 0.297} (41.510) &\bf 43.239 {\scriptsize $\pm$ 0.108} (43.327) \\
    0.70    & 37.487 {\scriptsize $\pm$ 0.285} (37.853) & 38.366 {\scriptsize $\pm$ 0.245} (38.631) &\bf 40.534 {\scriptsize $\pm$ 0.169} (40.834) \\
    0.75    & 33.977 {\scriptsize $\pm$ 0.232} (34.241) & 34.780 {\scriptsize $\pm$ 0.154} (34.865) &\bf 36.845 {\scriptsize $\pm$ 0.164} (37.123) \\
    0.80    & 30.191 {\scriptsize $\pm$ 0.158} (30.434) & 30.856 {\scriptsize $\pm$ 0.197} (30.874) &\bf 32.678 {\scriptsize $\pm$ 0.118} (32.772) \\
    0.85    & 25.119 {\scriptsize $\pm$ 0.157} (25.299) & 25.820 {\scriptsize $\pm$ 0.155} (25.849) &\bf 27.690 {\scriptsize $\pm$ 0.126} (27.712) \\
    0.90    & 19.152 {\scriptsize $\pm$ 0.164} (19.157) & 19.569 {\scriptsize $\pm$ 0.232} (19.617) &\bf 21.375 {\scriptsize $\pm$ 0.152} (21.487) \\
    0.95    & 08.028 {\scriptsize $\pm$ 0.290} (07.847) & 08.274 {\scriptsize $\pm$ 0.429} (08.647) &\bf 09.440 {\scriptsize $\pm$ 0.243} (09.286) \\ \midrule
\rowcolor{Gray}
    Average & 33.077 {\scriptsize $\pm$ 0.186} (33.315) & 33.837 {\scriptsize $\pm$ 0.189} (34.080) &\bf 35.712 {\scriptsize $\pm$ 0.062} (35.808) \\
    \bottomrule
    \end{tabular}
\end{subtable}

\end{table*}

\begin{table*}[t]
    \small
    \centering
    \caption{\textbf{TSP with other localization algorithms (extended results).} Each experiment in Study 3 is repeated five times, and we report the the mean, standard deviation (std), and max values over those five runs. Each table entry is given by \textit{\textbf{mean $\pm$ std (max)}}.}
    \label{table:supp_mat_extended_ablation_study_3}

\begin{subtable}{\linewidth}
    \centering
    \vspace{-5pt}
    \caption{\bf TAL on ActivityNet using BMN with R(2+1)D-18.}
    \vspace{-5pt}
    \begin{tabular}{c|ccc}
    \toprule 
            & \multicolumn{3}{c}{Feature Pretraining} \\ 
    mAP@tIoU& TAC on Kinetics                           & TAC on ActivityNet                        & TSP on ActivityNet \\ \midrule
    0.50    & 49.798 {\scriptsize $\pm$ 0.253} (49.951) & 50.339 {\scriptsize $\pm$ 0.270} (50.775) &\bf 51.283 {\scriptsize $\pm$ 0.206} (51.228) \\
    0.55    & 47.127 {\scriptsize $\pm$ 0.253} (47.391) & 47.731 {\scriptsize $\pm$ 0.289} (48.239) &\bf 48.665 {\scriptsize $\pm$ 0.209} (48.712) \\
    0.60    & 44.230 {\scriptsize $\pm$ 0.341} (44.621) & 44.760 {\scriptsize $\pm$ 0.252} (45.173) &\bf 45.759 {\scriptsize $\pm$ 0.176} (45.741) \\
    0.65    & 41.588 {\scriptsize $\pm$ 0.280} (41.905) & 42.027 {\scriptsize $\pm$ 0.252} (42.471) &\bf 43.310 {\scriptsize $\pm$ 0.166} (43.319) \\
    0.70    & 38.727 {\scriptsize $\pm$ 0.343} (39.078) & 39.016 {\scriptsize $\pm$ 0.232} (39.427) &\bf 40.346 {\scriptsize $\pm$ 0.122} (40.442) \\
    0.75    & 35.020 {\scriptsize $\pm$ 0.351} (35.306) & 35.201 {\scriptsize $\pm$ 0.178} (35.397) &\bf 36.577 {\scriptsize $\pm$ 0.165} (36.782) \\
    0.80    & 31.149 {\scriptsize $\pm$ 0.313} (31.521) & 31.494 {\scriptsize $\pm$ 0.190} (31.542) &\bf 32.609 {\scriptsize $\pm$ 0.186} (32.803) \\
    0.85    & 25.893 {\scriptsize $\pm$ 0.227} (26.186) & 26.315 {\scriptsize $\pm$ 0.217} (26.394) &\bf 27.398 {\scriptsize $\pm$ 0.108} (27.333) \\
    0.90    & 19.568 {\scriptsize $\pm$ 0.208} (19.980) & 19.991 {\scriptsize $\pm$ 0.276} (20.070) &\bf 20.825 {\scriptsize $\pm$ 0.196} (20.813) \\
    0.95    & 07.651 {\scriptsize $\pm$ 1.041} (08.613) &\bf 08.614 {\scriptsize $\pm$ 0.531} (07.963) & 08.420 {\scriptsize $\pm$ 0.557} (09.504) \\ \midrule
\rowcolor{Gray}
    Average & 34.075 {\scriptsize $\pm$ 0.304} (34.455) & 34.549 {\scriptsize $\pm$ 0.169} (34.745) &\bf 35.519 {\scriptsize $\pm$ 0.129} (35.668) \\
    \bottomrule
    \end{tabular}
\end{subtable}

\end{table*}

\begin{table*}[t]
    \small
    \centering
    \caption{\textbf{TSP on different pretraining datasets (extended results).} Each experiment in Study 4 is repeated five times, and we report the the mean, standard deviation (std), and max values over those five runs. Each table entry is given by \textit{\textbf{mean $\pm$ std (max)}}.}
    \label{table:supp_mat_extended_ablation_study_4}

\begin{subtable}{\linewidth}
    \centering
    \vspace{-5pt}
    \caption{\bf TAL on THUMOS14 using P-GCN with R(2+1)D-34.}
    \vspace{-5pt}
    \begin{tabular}{c|cc|cc}
    \toprule 
            & \multicolumn{4}{c}{Feature Pretraining} \\ 
    mAP@tIoU& TAC on Kinetics                           & TSP on ActivityNet                        & TAC on THUMOS14                           & TSP on THUMOS14 \\ \midrule
    0.1     & 70.978 {\scriptsize $\pm$ 0.157} (71.215) & 72.202 {\scriptsize $\pm$ 0.115} (72.193) & 71.713 {\scriptsize $\pm$ 0.090} (71.611) &\bf 73.889 {\scriptsize $\pm$ 0.116} (74.023) \\
    0.2     & 68.720 {\scriptsize $\pm$ 0.115} (68.640) & 69.836 {\scriptsize $\pm$ 0.139} (69.739) & 69.416 {\scriptsize $\pm$ 0.071} (69.362) &\bf 72.172 {\scriptsize $\pm$ 0.139} (72.286) \\
    0.3     & 65.876 {\scriptsize $\pm$ 0.087} (65.867) & 65.412 {\scriptsize $\pm$ 0.178} (65.403) & 66.289 {\scriptsize $\pm$ 0.080} (66.418) &\bf 68.840 {\scriptsize $\pm$ 0.165} (69.057) \\
    0.4     & 60.043 {\scriptsize $\pm$ 0.114} (60.048) & 59.844 {\scriptsize $\pm$ 0.165} (59.979) & 60.228 {\scriptsize $\pm$ 0.165} (60.302) &\bf 63.290 {\scriptsize $\pm$ 0.175} (63.314) \\
\rowcolor{Gray}
    0.5     & 48.763 {\scriptsize $\pm$ 0.348} (49.007) & 50.331 {\scriptsize $\pm$ 0.394} (51.038) & 49.879 {\scriptsize $\pm$ 0.421} (50.028) &\bf 52.901 {\scriptsize $\pm$ 0.337} (53.545) \\
    0.6     & 36.313 {\scriptsize $\pm$ 0.519} (37.048) & 36.339 {\scriptsize $\pm$ 0.298} (36.732) & 36.744 {\scriptsize $\pm$ 0.309} (36.559) &\bf 40.092 {\scriptsize $\pm$ 0.295} (40.445) \\
    0.7     & 22.380 {\scriptsize $\pm$ 0.386} (22.892) & 22.191 {\scriptsize $\pm$ 0.255} (22.221) & 22.769 {\scriptsize $\pm$ 0.302} (23.327) &\bf 25.691 {\scriptsize $\pm$ 0.210} (26.009) \\
    0.8     & 09.325 {\scriptsize $\pm$ 0.199} (09.126) & 09.270 {\scriptsize $\pm$ 0.203} (09.285) & 09.508 {\scriptsize $\pm$ 0.157} (09.713) &\bf 10.619 {\scriptsize $\pm$ 0.229} (10.469) \\
    0.9     & 01.413 {\scriptsize $\pm$ 0.071} (01.409) & 01.399 {\scriptsize $\pm$ 0.067} (01.393) & 01.435 {\scriptsize $\pm$ 0.103} (01.484) &\bf 01.615 {\scriptsize $\pm$ 0.071} (01.674) \\
    \bottomrule
    \end{tabular}
\end{subtable}

\begin{subtable}{\linewidth}
    \centering
    \vspace{5pt}
    \caption{\bf TAL on THUMOS14 using G-TAD with R(2+1)D-34.}
    \vspace{-5pt}
    \begin{tabular}{c|cc|cc}
    \toprule 
            & \multicolumn{4}{c}{Feature Pretraining} \\ 
    mAP@tIoU& TAC on Kinetics                           & TSP on ActivityNet                        & TAC on THUMOS14                           & TSP on THUMOS14 \\ \midrule
    0.1     & 58.311 {\scriptsize $\pm$ 0.553} (58.934) & 60.747 {\scriptsize $\pm$ 1.114} (62.106) & 59.747 {\scriptsize $\pm$ 0.655} (60.546) &\bf 67.605 {\scriptsize $\pm$ 1.096} (68.498) \\
    0.2     & 54.909 {\scriptsize $\pm$ 0.383} (55.446) & 57.173 {\scriptsize $\pm$ 1.144} (58.991) & 56.756 {\scriptsize $\pm$ 0.662} (57.738) &\bf 64.542 {\scriptsize $\pm$ 1.106} (65.279) \\
    0.3     & 49.728 {\scriptsize $\pm$ 0.685} (50.590) & 51.622 {\scriptsize $\pm$ 1.136} (53.449) & 51.202 {\scriptsize $\pm$ 0.717} (52.608) &\bf 58.205 {\scriptsize $\pm$ 1.236} (59.628) \\
    0.4     & 42.405 {\scriptsize $\pm$ 0.591} (43.232) & 43.945 {\scriptsize $\pm$ 1.163} (45.924) & 43.999 {\scriptsize $\pm$ 0.708} (45.538) &\bf 50.853 {\scriptsize $\pm$ 1.243} (51.987) \\
\rowcolor{Gray}
    0.5     & 33.255 {\scriptsize $\pm$ 0.760} (34.521) & 35.089 {\scriptsize $\pm$ 1.066} (37.034) & 34.797 {\scriptsize $\pm$ 0.558} (35.823) &\bf 41.500 {\scriptsize $\pm$ 1.118} (43.232) \\
    0.6     & 23.618 {\scriptsize $\pm$ 0.615} (24.080) & 24.865 {\scriptsize $\pm$ 1.027} (26.734) & 25.024 {\scriptsize $\pm$ 0.747} (26.194) &\bf 30.196 {\scriptsize $\pm$ 1.422} (32.201) \\
    0.7     & 14.467 {\scriptsize $\pm$ 0.918} (15.467) & 14.771 {\scriptsize $\pm$ 0.789} (16.128) & 15.536 {\scriptsize $\pm$ 0.557} (15.565) &\bf 18.446 {\scriptsize $\pm$ 1.318} (21.052) \\
    0.8     & 06.763 {\scriptsize $\pm$ 0.694} (07.254) & 06.479 {\scriptsize $\pm$ 0.498} (07.403) & 07.264 {\scriptsize $\pm$ 0.401} (07.231) &\bf 08.836 {\scriptsize $\pm$ 0.873} (10.592) \\
    0.9     & 01.224 {\scriptsize $\pm$ 0.157} (01.313) & 01.086 {\scriptsize $\pm$ 0.155} (01.351) & 01.271 {\scriptsize $\pm$ 0.113} (01.355) &\bf 01.491 {\scriptsize $\pm$ 0.133} (01.721) \\
    \bottomrule
    \end{tabular}
\end{subtable}

\end{table*}

\section{Extended State-of-the-Art Comparison}

\begin{table*}[t]
    \small
    \centering
    \caption{\textbf{SOTA comparison for TAL on ActivityNet (extended results).} We use G-TAD as the algorithms atop our features. TSP achieves SOTA performance.}
    \begin{tabular}{l|cccg}
    \toprule
Method                       &   0.5   &   0.75  &   0.95  &   Avg. \\ \midrule
R-C3D~\cite{xu_iccv_2017}    &   26.80 &   --    &   --    &   --    \\
TAL-Net~\cite{chao_cvpr_2018}&   38.23 &   18.30 &   1.30  &   20.22 \\
SCC~\cite{caba2017scc}       &   40.00 &   17.90 &   4.70  &   21.70 \\
TCN~\cite{dai2017temporal}   &   37.49 &   23.47 &   4.47  &   23.58 \\
CDC~\cite{shou_cvpr_2017}    &   45.30 &   26.00 &   0.20  &   23.80 \\
BSN~\cite{lin_eccv_2018}     &   46.45 &   29.96 &   8.02  &   30.03 \\
Zhao~\etal~\cite{zhao2020bottom}&43.47 &   33.91 &   9.21  &   30.12 \\
C-TCN~\cite{li2020deep}      &   47.60 &   31.90 &   6.20  &   31.10 \\
P-GCN~\cite{Zeng_2019_ICCV}  &   48.26 &   33.16 &   3.27  &   31.11 \\
BMN~\cite{Lin_2019_ICCV}     &   50.07 &   34.78 &   8.29  &   33.85 \\
GTAN~\cite{long2019gaussian} &   52.61 &   34.14 &   8.91  &   34.31 \\
PBRNet~\cite{liu2020progressive} &\bf53.96 &   34.97 &   8.98  &   35.01 \\
\midrule
G-TAD~\cite{xu2020gtad}      &   50.36 &   34.60 &   9.02  &   34.09 \\
\bf TSP (ours)               &   51.26 &\bf37.12 &\bf9.29  &\bf35.81 \\
\bottomrule
    \end{tabular}
    \label{table:supp_mat_sota_tal_anet}
\end{table*}

\begin{table*}[t]
    \small
    \centering
    \caption{\textbf{SOTA comparison for TAL on THUMOS14 (extended results).} We use P-GCN as the algorithms atop our features. TSP achieves SOTA performance.}
    \begin{tabular}{l|ccccgcccc}
    \toprule
Method                       &   0.1  &   0.2  &   0.3  &   0.4  &   0.5  &   0.6  &   0.7  &   0.8  &   0.9  \\
\midrule
Hou~\etal~\cite{hou2017real} &   51.3 &   --   &   43.7 &   --   &   22.0 &   --   &   --   &   --   &   --   \\ 
SST~\cite{buch_cvpr_2017}    &   --   &   --   &   37.8 &   --   &   23.0 &   --   &   --   &   --   &   --   \\ 
CDC~\cite{shou_cvpr_2017}    &   --   &   --   &   40.1 &   29.4 &   23.3 &   13.1 &   7.9  &   --   &   --   \\ 
TCN~\cite{dai2017temporal}   &   --   &   --   &   --   &   33.3 &   25.6 &   15.9 &   9.0  &   --   &   --   \\ 
TURN-TAP~\cite{gao_iccv_2017}&   54.0 &   50.9 &   44.1 &   34.9 &   25.6 &   --   &   --   &   --   &   --   \\ 
R-C3D~\cite{xu_iccv_2017}    &   54.5 &   51.5 &   44.8 &   35.6 &   28.9 &   --   &   --   &   --   &   --   \\
SS-TAD~\cite{buch2019end}    &   --   &   --   &   45.7 &   --   &   29.2 &   --   &   9.6  &   --   &   --   \\ 
SSN~\cite{zhao_iccv_2017}    &   66.0 &   59.4 &   51.9 &   41.0 &   29.8 &   --   &   --   &   --   &   --   \\ 
CTAP~\cite{gao_eccv_2018}    &   --   &   --   &   --   &   --   &   29.9 &   --   &   --   &   --   &   --   \\
Action Search~\cite{alwassel_2018_actionsearch}&--&--&51.8& 42.4 &   30.8 &   20.2 &   11.1 &   --   &   --   \\
CBR~\cite{cbr}               &   60.1 &   56.7 &   50.1 &   41.3 &   31.0 &   19.1 &   9.9  &   --   &   --   \\
ETP~\cite{qiu2018precise}    &   --   &   --   &   48.2 &   42.4 &   34.2 &   23.4 &   13.9 &   --   &   --   \\
BSN~\cite{lin_eccv_2018}     &   --   &   --   &   53.5 &   45.0 &   36.9 &   28.4 &   20.0 &   --   &   --   \\
MGG\cite{mgg}                &   --   &   --   &   53.9 &   46.8 &   37.4 &   29.5 &   21.3 &   --   &   --   \\
GTAN~\cite{long2019gaussian} &   --   &   --   &   57.8 &   47.2 &   38.8 &   --   &   --   &   --   &   --   \\
BMN~\cite{Lin_2019_ICCV}     &   --   &   --   &   56.0 &   47.4 &   38.8 &   29.7 &   20.5 &   --   &   --   \\
DBG~\cite{dbg}               &   --   &   --   &   57.8 &   49.4 &   39.8 &   30.2 &   21.7 &   --   &   --   \\ 
CMS-RC3D~\cite{CMS-RC3D}     &   61.6 &   59.3 &   54.7 &   48.2 &   40.0 &   --   &   --   &   --   &   --   \\
G-TAD~\cite{xu2020gtad}      &   --   &   --   &   54.5 &   47.6 &   40.2 &   30.8 &   23.4 &   --   &   --   \\ 
TAL-Net~\cite{chao_cvpr_2018}&   59.8 &   57.1 &   53.2 &   48.5 &   42.8 &   33.8 &   20.8 &   --   &   --   \\ 
Zhao~\etal~\cite{zhao2020bottom}&--   &   --   &   53.9 &   50.7 &   45.4 &   38.0 &   28.5 &   --   &   --   \\ 
PBRNet~\cite{liu2020progressive} &   --   &   --   &   58.5 &   54.6 &   51.3 &   41.8 &\bf29.5 &   --   &   --   \\
C-TCN~\cite{li2020deep}      &   72.2 &   71.4 &   68.0 &   62.3 &   52.1 &   --   &   --   &   --   &   --   \\
TSA-Net~\cite{TSA_Net}       &   --   &   --   &   65.6 &   61.4 &   53.0 &\bf42.4 &   28.8 &   --   &   --   \\
\midrule
P-GCN~\cite{Zeng_2019_ICCV}  &   69.5 &   67.8 &   63.6 &   57.8 &   49.1 &   --   &   --   &   --   &   --   \\ 
\bf TSP (ours)               &\bf74.0 &\bf72.3 &\bf69.1 &\bf63.3 &\bf53.5 &   40.4 &   26.0 &\bf10.5 &\bf1.7  \\ 
\bottomrule
    \end{tabular}
    \label{table:supp_mat_sota_tal_thumos}
\end{table*}

\begin{table*}[t]
    \small
    \centering
    \caption{\textbf{SOTA comparison for Dense-Captioning on ActivityNet Captions (extended results).} We use BMT as the algorithms atop our features. TSP achieves SOTA performance in terms of average BLEU and is competitive in terms of average METEOR. The best numbers are highlighted in bold and the second best is underlined.}
    \begin{tabular}{l|ccg|ccg}
\toprule
                    & \multicolumn{3}{c|}{Ground Truth Proposals} & \multicolumn{3}{c}{Learned Proposals} \\ 
Method              &\footnotesize BLEU@3 &\footnotesize BLEU@4 & METEOR &\footnotesize BLEU@3 &\footnotesize BLEU@4 & METEOR \\
\midrule
Rahman~\etal~\cite{rahman2019watch}&   3.04 &   1.46 &    7.23 &   1.85 &   0.90 &   4.93 \\
Krishna~\etal~\cite{activitynet_captions_dataset}&   4.09 &   1.60 &    8.88 &   1.90 &   0.71 &   5.69 \\ 
Bi-SST~\cite{Wang_2018_CVPR}&   --   &   --   &   10.89 &   2.27 &   1.13 &   6.10 \\
Masked Transformer~\cite{Zhou_2018_CVPR}&\bf5.76 &\bf2.71 &   11.16 &   2.91 &   1.44 &   6.91 \\
DVC~\cite{Li_dvc}   &   4.55 &   1.62 &   10.33 &   2.27 &   0.73 &   6.93 \\ 
MFT~\cite{mft}      &   --   &   --   &   --    &   2.82 &   1.24 &   7.08 \\ 
MDVC~\cite{mdvc}    &   4.52 &   1.98 &   11.07 &   2.53 &   1.01 &   7.46 \\ 
SDVC~\cite{sdvc}    &   4.41 &   1.28 &\bf13.07 &   2.94 &   0.93 &\bf8.82 \\
\midrule
BMT~\cite{bmt}      &   4.63 &\underline{1.99} &   10.90 &\underline{3.84} &\underline{1.88} &   8.44 \\
\bf TSP (ours)      &\underline{4.76} &\underline{1.99} &\underline{11.31} &\bf4.16 &\bf2.02 &\underline{8.75} \\
\bottomrule 
    \end{tabular}
    \label{table:supp_mat_sota_captioning}
\end{table*}

\begin{table*}[t]
    \small
    \centering
    \tabcolsep=0.1cm
    \caption{\textbf{SOTA comparison for Proposals on ActivityNet (extended results)}. We use BMN atop our features. TSP significantly improves over BMN original performance, and is competitive with SOTA.}
    \begin{tabular}{l|ccccccc|cc}
    \toprule
 Method  & \cite{gao_eccv_2018} & \cite{lin_eccv_2018} & \cite{mgg} & \cite{zhao2020bottom} & \cite{bcgcn} & \cite{dbg} & \cite{gao2020accurate} & BMN~\cite{Lin_2019_ICCV}  &\bf TSP  \\ \midrule
 AR@100  & 73.17                & 74.16                & 74.54      & 75.27                 & 76.73        & 76.65      &\bf78.63                & 75.01                     &   76.63 \\
\rowcolor{Gray}
 AUC & 65.72                & 66.17                & 66.43      & 66.51                 & 68.05        & 68.23      &\bf69.93                & 67.10                     &   69.04 \\
    \bottomrule
    \end{tabular}
    \label{table:supp_mat_sota_proposals}
\end{table*}

Tables~\ref{table:supp_mat_sota_tal_anet}, \ref{table:supp_mat_sota_tal_thumos}, \ref{table:supp_mat_sota_captioning}, and \ref{table:supp_mat_sota_proposals} present extended SOTA comparison with more methods and all tIoUs for TAL on ActivityNet, TAL on THUMOS14, Dense-Captioning on ActivityNet Captions, and Proposals on ActivityNet, respectively. For the Dense-Captioning task, we report additional results for captioning ground truth proposals.

\section{Extended Feature Analysis Study}

\begin{figure*}[t]
    \includegraphics[width=0.121\linewidth]{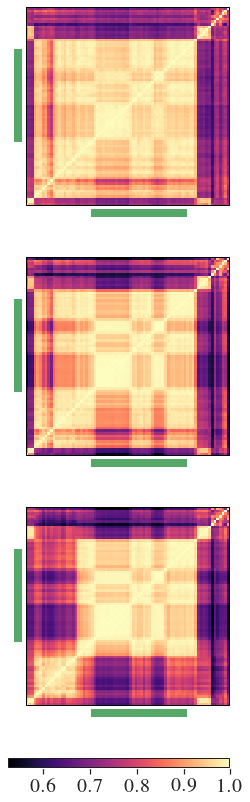}
    \includegraphics[width=0.121\linewidth]{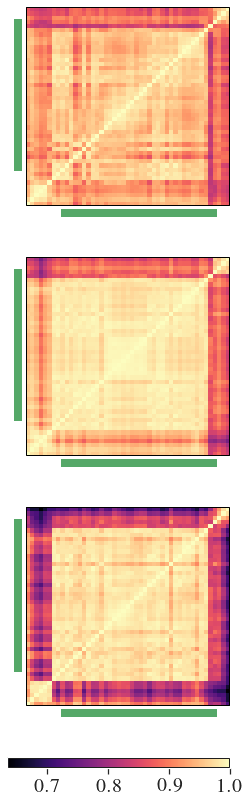}    
    \includegraphics[width=0.121\linewidth]{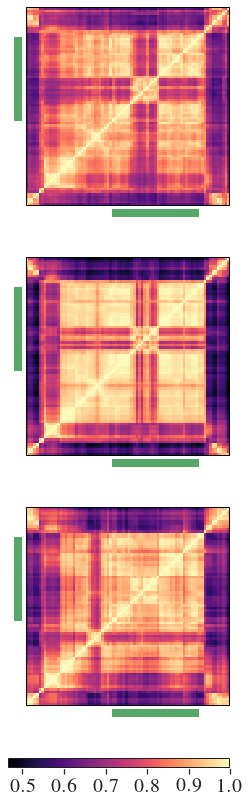}
    \includegraphics[width=0.121\linewidth]{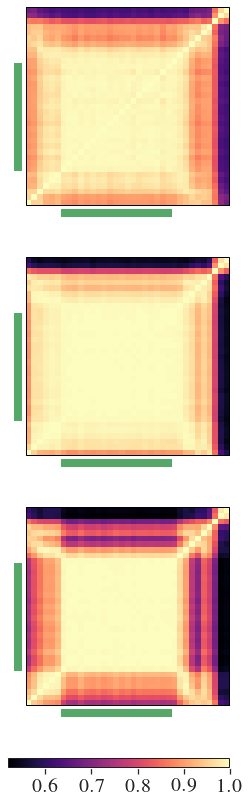}
    \includegraphics[width=0.121\linewidth]{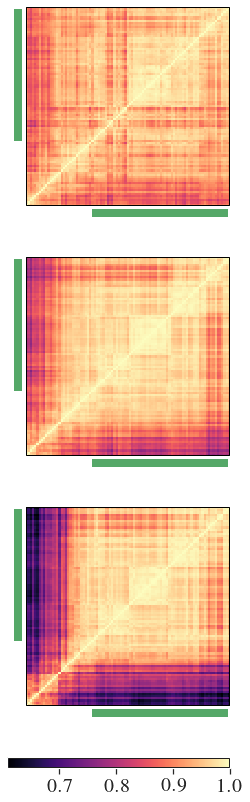}
    \includegraphics[width=0.121\linewidth]{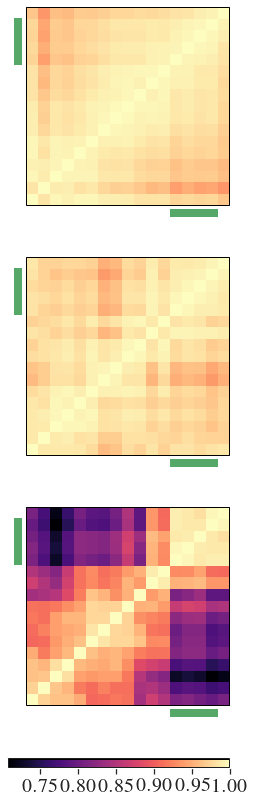}
    \includegraphics[width=0.121\linewidth]{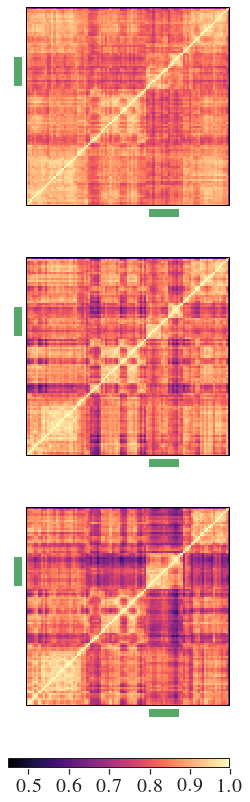}
    \includegraphics[width=0.121\linewidth]{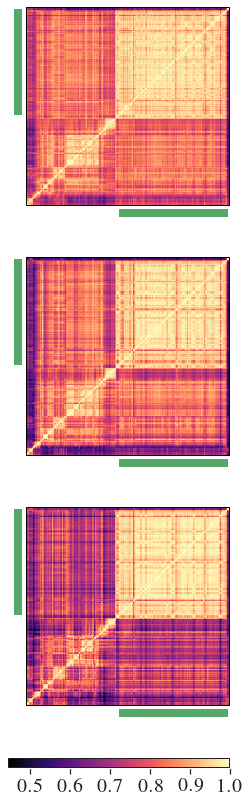}\\
    \includegraphics[width=0.121\linewidth]{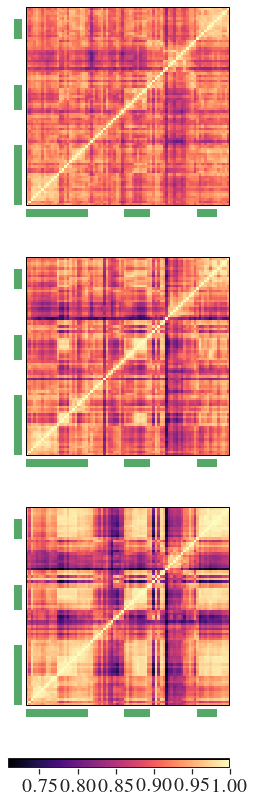}
    \includegraphics[width=0.121\linewidth]{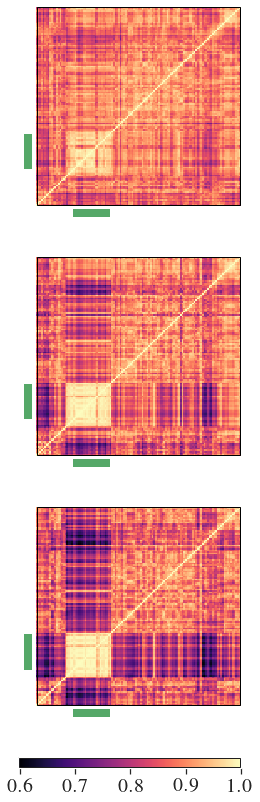}
    \includegraphics[width=0.121\linewidth]{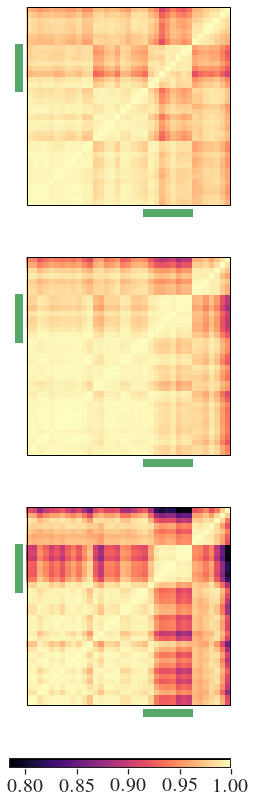}
    \includegraphics[width=0.121\linewidth]{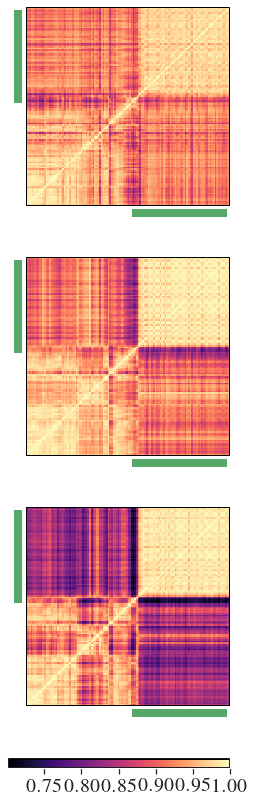}
    \includegraphics[width=0.121\linewidth]{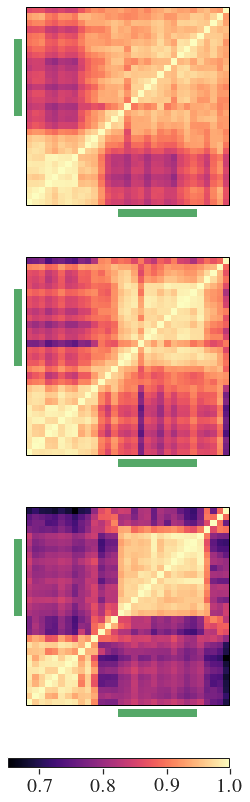}
    \includegraphics[width=0.121\linewidth]{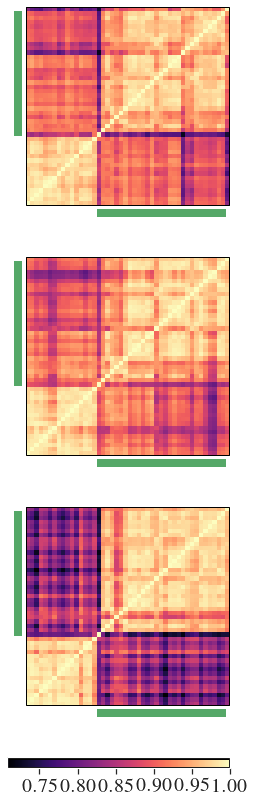}
    \includegraphics[width=0.121\linewidth]{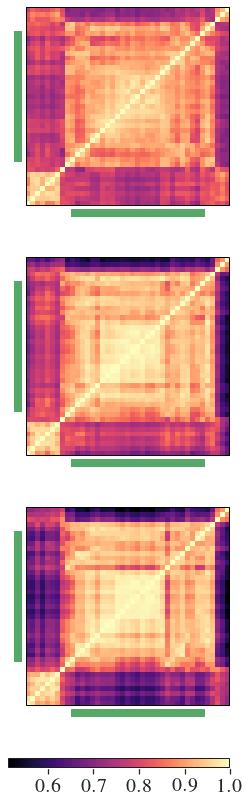}
    \includegraphics[width=0.121\linewidth]{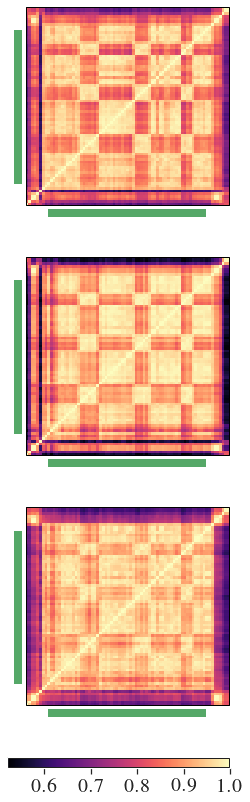}
    \\
    \vspace{-5pt}
    \caption{\textbf{Feature similarity (extended results)}. Each column  (set of three matrices) shows the similarity matrices of one video using \textit{TAC on Kinetics} (top), \textit{TAC on ActivityNet} (middle), and TSP on ActivityNet (bottom) features. The green lines next to each matrix represent the temporal extent of ground truth actions. Better viewed in color.}
    \vspace{-5pt}
    \label{fig:supp_mat_cosine_similarity}
\end{figure*}

Here, we extended the feature similarity study to compare with \textit{TAC on ActivityNet}. Figure~\ref{fig:supp_mat_cosine_similarity} provides more examples comparing TSP features with those of \textit{TAC on Kinetics} and \textit{TAC on ActivityNet}. Not only does TSP show better temporal sensitivity compared to \textit{TAC on Kinetics} (as we have shown in the main paper), but it also presents a better distinguishing of background \vs foreground representation compared to \textit{TAC on ActivityNet}. 

\end{document}